\documentclass[10pt,twocolumn,letterpaper]{article}

\usepackage[pagenumbers]{cvpr} 

\usepackage[dvipsnames]{xcolor}
\newcommand{\red}[1]{{\color{red}#1}}

\usepackage[pagebackref,breaklinks,colorlinks,citecolor=teal]{hyperref}

\usepackage{xcolor}
\definecolor{olivegreen}{HTML}{3C8031}

\newcommand\blue[1]{{\textcolor{blue}{#1}}}

\definecolor{darkergreen}{RGB}{21, 152, 56}
\newcommand\greenp[1]{\textcolor{darkergreen}{(#1)}}
\newcommand\greenpscript[1]{\scriptsize\greenp{#1}} 

\usepackage{booktabs}
\usepackage{adjustbox}
\usepackage{subcaption}
\usepackage{bbm}
\usepackage{multirow}

\title{Towards Label-Efficient Human Matting: \\A Simple Baseline for Weakly Semi-Supervised Trimap-Free Human Matting}

\author{
Beomyoung Kim$^{1,2}$\hspace{1.5em}Myeong Yeon Yi$^{1}$\hspace{1.5em}Joonsang Yu$^{1}$\hspace{1.5em}Young Joon Yoo$^{3}$\hspace{1.5em}Sung Ju Hwang$^{2}$\\ \\
{NAVER Cloud, ImageVision$^1$\hspace{3em}KAIST$^2$\hspace{3em}Chung-Ang University$^3$}\\
}

\begin{document}
\maketitle

\begin{abstract}
This paper presents a new practical training method for human matting, which demands delicate pixel-level human region identification and significantly laborious annotations.
To reduce the annotation cost, most existing matting approaches often rely on image synthesis to augment the dataset.
However, the unnaturalness of synthesized training images brings in a new domain generalization challenge for natural images.
To address this challenge, we introduce a new learning paradigm, weakly semi-supervised human matting (WSSHM), which leverages a small amount of expensive matte labels and a large amount of budget-friendly segmentation labels, to save the annotation cost and resolve the domain generalization problem.
To achieve the goal of WSSHM, we propose a simple and effective training method, named Matte Label Blending (MLB), that selectively guides only the beneficial knowledge of the segmentation and matte data to the matting model.
Extensive experiments with our detailed analysis demonstrate our method can substantially improve the robustness of the matting model using a few matte data and numerous segmentation data. Our training method is also easily applicable to real-time models, achieving competitive accuracy with breakneck inference speed (328 FPS on NVIDIA V100 GPU). The implementation code is available at \url{https://github.com/clovaai/WSSHM}.
\end{abstract}

\section{Introduction}
\label{sec:intro}

\begin{figure}[t]
    \centering
    \begin{subfigure}[b]{\linewidth}
        \centering
        \includegraphics[width=\linewidth]{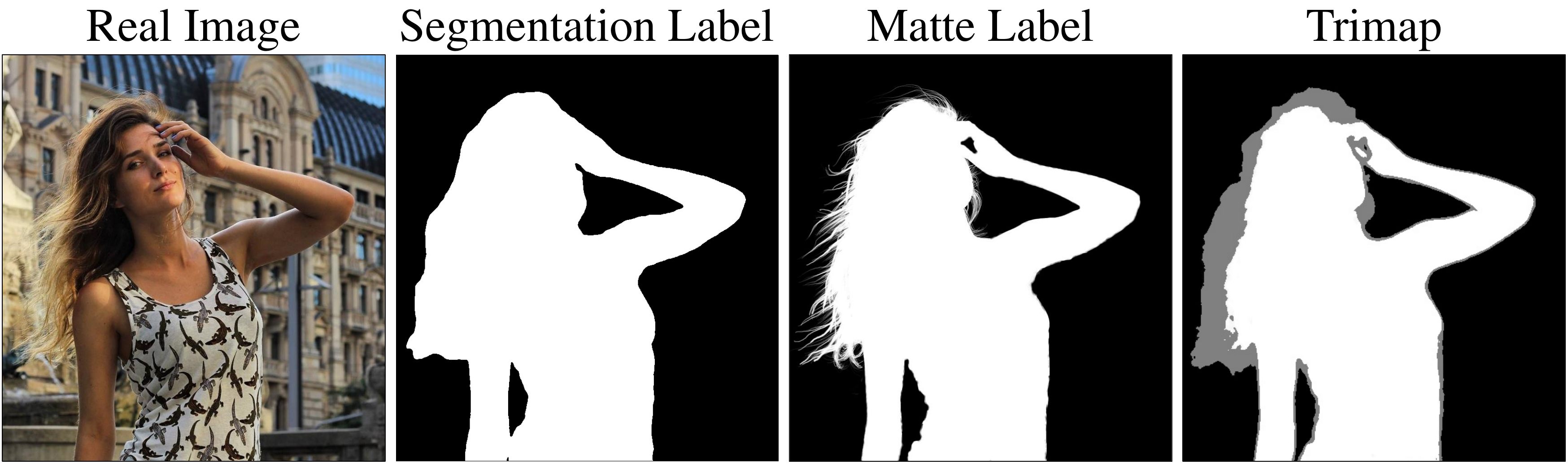} 
        \caption{label definition}
        \label{fig:label_definition}  
        \vspace{3mm}
    \end{subfigure}
    \begin{subfigure}[b]{\linewidth}
        \centering
        \includegraphics[width=\linewidth]{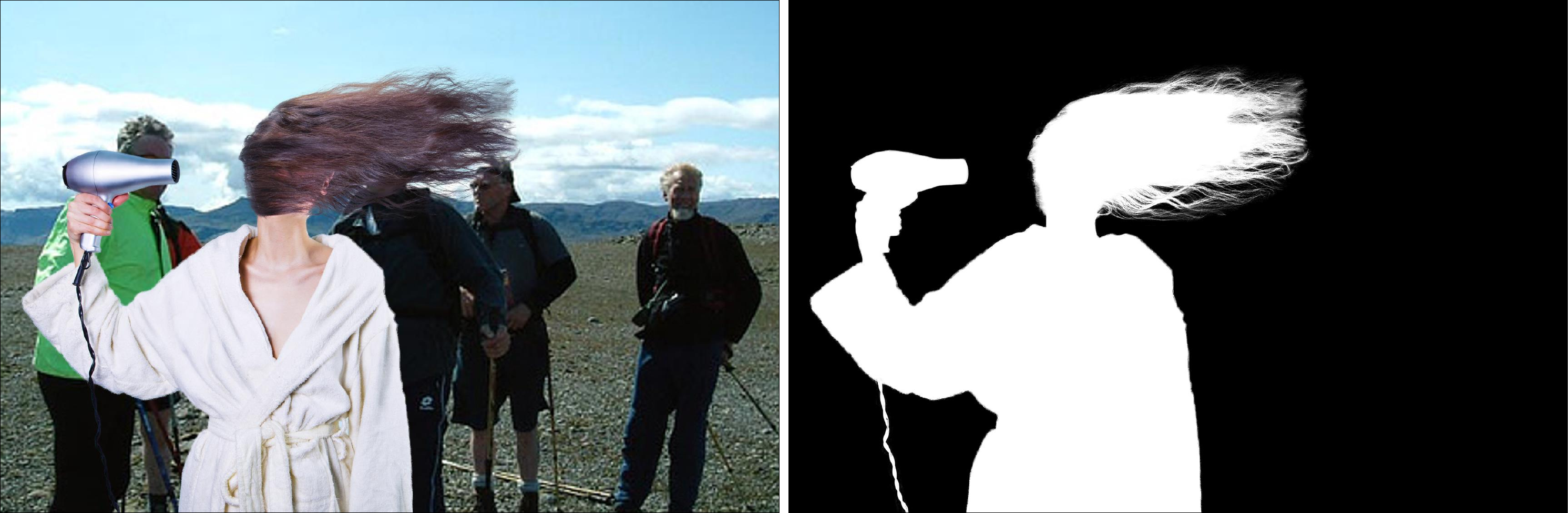} 
        \caption{synthetic matte data}
        \label{fig:matte_synthetic} 
    \end{subfigure}
    \caption{
    (a) The forms of segmentation label, matte label, and trimap.
    The matte label requires much more sophisticated annotations than the segmentation label.
    The gray-colored region in the trimap is the unknown region where details need to be estimated.
    (b) An example of synthetic matte data that is unnatural and lacks real-world context.
    }
    \label{fig:form_seg_and_matte_labels}
\end{figure}

Human matting aims to recognize the precise per-pixel opacity of human regions with significantly more elaborated details than human segmentation, and it is widely used for many applications, such as image editing and background replacement.
However, obtaining matte labels is extremely costly and time-consuming since the annotations must include precise boundaries of every detail, including small hair strands, as shown in Figure \ref{fig:label_definition}, which can be a daunting task for human annotators.

Unfortunately, publicly available matting datasets are extremely limited due to the high cost of matte annotations.
For instance, Human-2K~\cite{(human2k)liu2021tripartite}, Distinctions-646~\cite{(distinctions)(HAttMatting)qiao2020attention}, and DIM~\cite{(dim)xu2017dim} consist of 2000, 362, and 202 images, respectively,
Moreover, most of the datasets provide only foreground images.
To mitigate the scarcity of labeled data, prior works~\cite{(shm)chen2018semantic,(PPM)ke2022modnet,(human2k)liu2021tripartite,(matteformer)park2022matteformer,(distinctions)(HAttMatting)qiao2020attention,(RWP)(MGMatte)yu2021mask,(LFM)zhang2019late,(dim)xu2017dim} utilize synthetic images to augment the datasets, as shown in \figurename~\ref{fig:matte_synthetic}.
However, synthetic images lack the natural contextual compositions of real-world images, causing a domain generalization problem.

\begin{figure}[t]
    \centering
    \includegraphics[width=\linewidth]{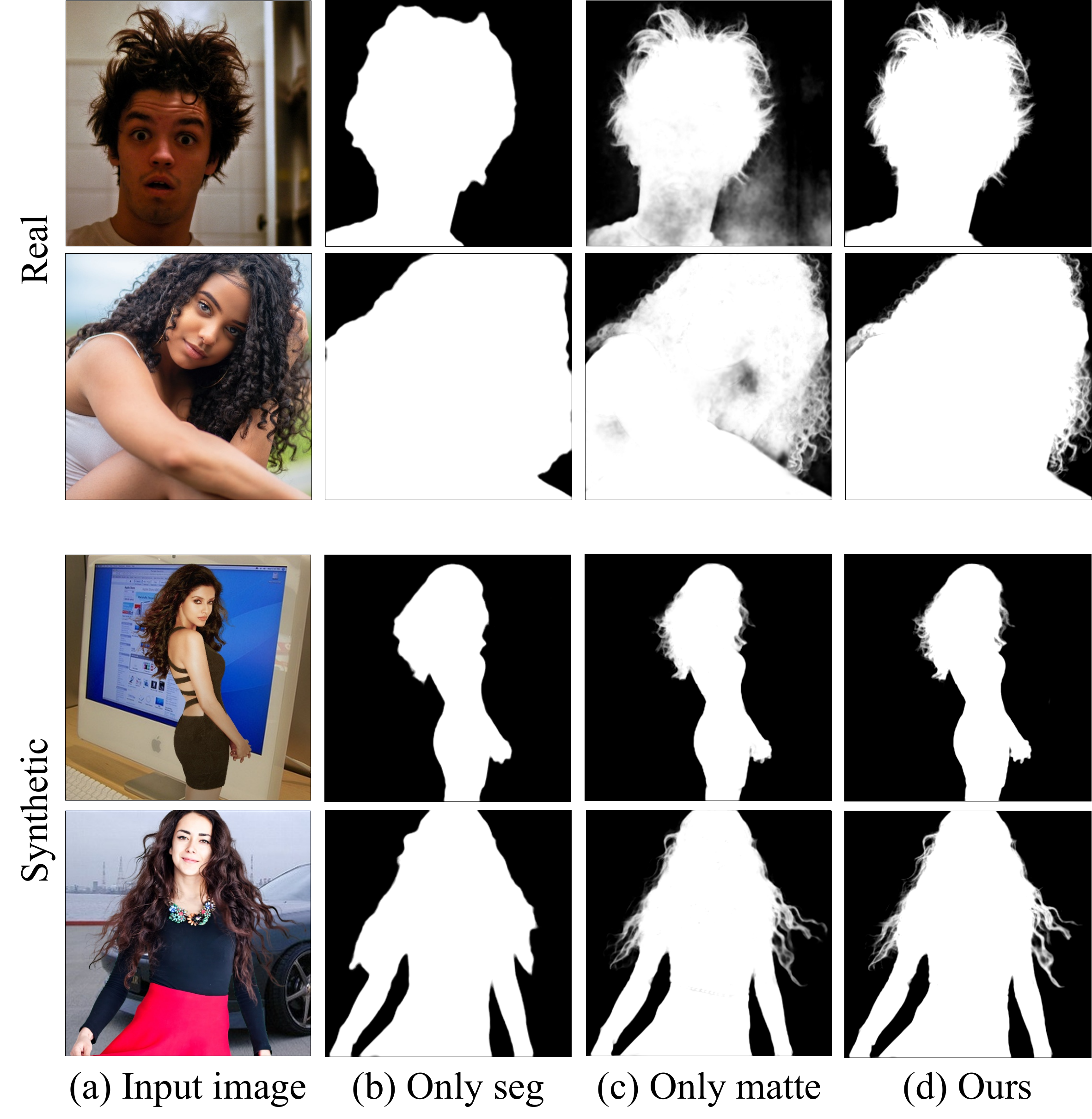}
    \caption{
    The qualitative comparisons of the models trained with (b) only segmentation data, (c) only matte data, (d) and both segmentation and matte data with our training method. 
    The first and second rows are the results on natural (real-world) images, and the third and fourth rows are the results on synthetic images. 
    }
    \label{fig:sample_domain_generalization}
\end{figure}

To circumvent this problem, previous works~\cite{(shm)chen2018semantic,cho2016natural,(human2k)liu2021tripartite,lu2019indices,(matteformer)park2022matteformer,shen2016deep,tang2019learning,(dim)xu2017dim} exploit an extra input source called trimap, which contains explicit foreground, background, and unknown region information, as shown in \figurename~\ref{fig:label_definition}.
By using the trimap as a strong prior, the solution space is reduced to the unknown (boundary) region, which is the gray-colored region in \figurename~\ref{fig:label_definition}, leading to significant performance improvement on natural images even when trained with only synthetic data. 
However, trimap-based methods are not suitable for real-world scenarios such as real-time and non-interactive applications.
To facilitate real-world applications, trimap-free approaches, which take only an image as input, have attracted more attention recently.
However, when trained with only synthetic data, they often fail to generalize to natural images, as shown in \figurename~\ref{fig:sample_domain_generalization}, since the absence of the trimap makes the model more vulnerable to the domain generalization problem.
Building more matte data for natural images is the obvious solution, but it is not practical due to the tremendous annotation cost.

\begin{figure}[t]
    \centering
    \includegraphics[width=0.92\linewidth]{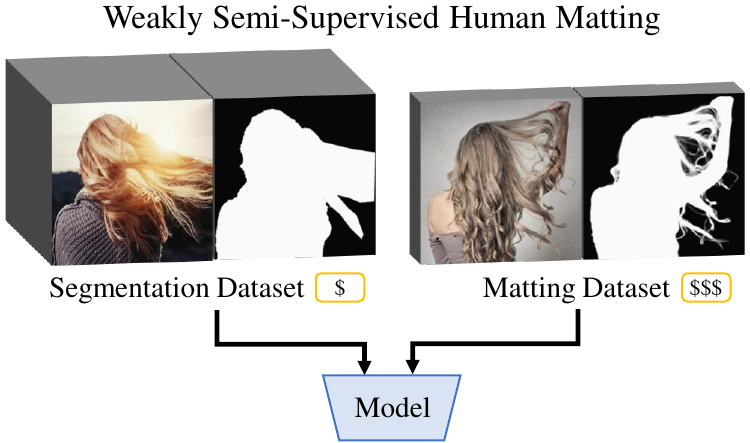}
    \caption{Illustration for the definition of Weakly Semi-Supervised Human Matting (WSSHM). 
    The goal of the WSSHM is to train the trimap-free matting model using a small amount of expensive matte data and a large amount of economic segmentation data.}
    \label{fig:task_definition}
\end{figure}

To address these challenges, we present a new learning paradigm for human matting, named \textbf{Weakly Semi-Supervised trimap-free Human Matting (WSSHM)}, to reduce the costly annotation of matte labels while improving the robustness of the trimap-free model to natural images.
The WSSHM leverages a small amount of expensive matte data and a large amount of budget-friendly segmentation data, considering the segmentation label as a weak label and the matte label as a strong label, as illustrated in Figure \ref{fig:task_definition}.
Although segmentation labels lack detailed human boundary information as shown in Figure \ref{fig:label_definition}, they are significantly cheaper to obtain than matte labels and are publicly available in massive quantities for natural images.
In this paper, we consider that the matting dataset consists of only synthetic images and the segmentation dataset consists of only natural images, which is a realistic setting in matting research fields.

To achieve the goal of WSSHM, we introduce a baseline training strategy named \textbf{Matte Label Blending (MLB)}, which aims to maximize the synergy between segmentation and matte data in a simple and effective way.
The MLB selectively leverages beneficial information from both segmentation and matte labels.
Namely, segmentation data is employed to enhance the robustness of the model to natural images, and synthetic matte data is used to improve the boundary detail representation of the model.

To verify the effectiveness of the proposed method under the WSSHM setting, we precisely analyze the effect of the amount of segmentation and matte labels with extensive experiments.
The results demonstrate that \textbf{(1)} leveraging the cost-efficient segmentation data can dramatically improve the domain generalization of the model, \textbf{(2)} using a small amount of matte data significantly enhances the boundary detail representation of the model, \textbf{(3)} the proposed method can achieve promising matte results on natural images without any matting dataset consisting of natural images, \textbf{(4)} our training method can be easily applicable to existing models.
In summary, our contributions are as follows:
\begin{itemize}
    \item We introduce a new learning paradigm for human matting, Weakly Semi-supervised Trimap-free Human Matting (WSSHM) to reduce annotation costs and improve domain generalization of the model to natural images. 
    \item We propose an effective baseline method for WSSHM, named Matte Label Blending (MLB), which selectively exploits advantageous information from the matte and segmentation labels.
\end{itemize}

\begin{figure*}[t]
    \centering
    \includegraphics[width=\linewidth]{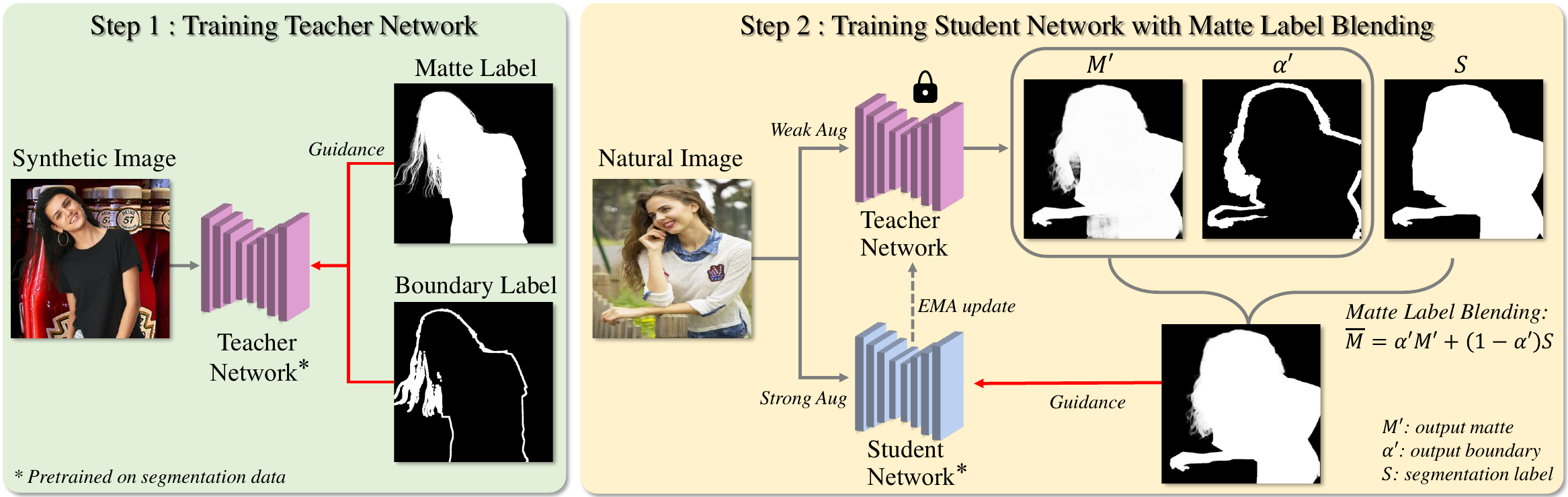}
    \caption{\textbf{Overview of the proposed training pipeline} encompassing the two-step process involving (1st step) teacher network training with synthetic matte data and (2nd step) student network training with natural segmentation data, along with the integration of Matte Label Blending (MLB) mechanism to combine boundary detail representation and domain generalization to achieve the goal of WSSHM.}
    \label{fig:overview}
\end{figure*}

\section{Related Work}
\label{sec:related_work}

\subsection{Human Matting}
Compared to human segmentation, human matting requires detailed pixel-level estimation of the opacity of the human region. 
Due to the challenging nature of the matting task, most existing methods~\cite{(shm)chen2018semantic,cho2016natural,(human2k)liu2021tripartite,lu2019indices,shen2016deep,tang2019learning} have employed a trimap as an auxiliary input source to focus on only the boundary region and achieve promising matte results. 
DIM~\cite{(dim)xu2017dim} proposed an encoder-decoder network with a sequential refinement stage.
MatteFormer~\cite{(matteformer)park2022matteformer} adopts the vision transformer architecture~\cite{(vit)dosovitskiy2020image} and leverages the trimap as the prior token of the transformer.
Additionally, some works~\cite{(RWP)(MGMatte)yu2021mask, park2023mask} utilize a coarse segmentation mask as a form of trimap.
However, the necessity of obtaining the trimap (or segmentation mask) in the input space demands extra annotations or user interaction, limiting its practical applicability in real-world scenarios and real-time applications.

For the better real-world applicability, trimap-free methods~\cite{(PPM)ke2022modnet,(boosting)liu2020boosting, (distinctions)(HAttMatting)qiao2020attention,(LFM)zhang2019late,(fdmpa)zhu2017fast} have been proposed.
LFM~\cite{(LFM)zhang2019late} proposed a dual-decoder network for foreground and background identification with a fusion module.
MODNet~\cite{(PPM)ke2022modnet} presented a real-time portrait matting model that consists of three interdependent branches with a series of sub-objectives.
However, due to the absence of the trimap, their performance still lags behind trimap-based methods.

BSHM~\cite{(boosting)liu2020boosting} incorporates segmentation labels to boost matting performance, but they require three sequential sub-networks and lack detailed analysis of performance variation with the amount of segmentation and matte data. 
In contrast, our training method is more straightforward and flexibly applicable to existing real-time models, and we provide in-depth analyses of this synergy between the amount of segmentation and matte data.

\subsection{Weakly Semi-Supervised Learning}
In the field of object detection, there exist several approaches~\cite{(point_detr)chen2021points,fang2021wssod,(wssod)yan2017weakly,(group_rcnn)zhang2022group,(pointwssis)kim2023devil} to minimize annotation costs and achieve reasonably strong performance by utilizing a few amounts of fully-labeled data ($i.e.,$ bounding boxes) and numerous weakly-labeled data ($e.g.,$ image-level or point labels), known as weakly semi-supervised object detection.
Inspired by this learning paradigm, we present weakly semi-supervised human matting (WSSHM), which considers the segmentation label as the weak label and the matte label as the strong label, to reduce annotation costs while achieving robust matting results.

\section{Proposed Method}

\subsection{Problem Setting}
Due to the high cost of annotating matte labels and the limited availability of public matting datasets, existing human matting methods~\cite{(PPM)ke2022modnet, (matteformer)park2022matteformer,(distinctions)(HAttMatting)qiao2020attention, (RWP)(MGMatte)yu2021mask, (LFM)zhang2019late} heavily rely on synthetic datasets. 
However, models trained solely on synthetic data often fail to generalize to natural images due to the lack of natural compositions in synthetic images (Figure~\ref{fig:sample_domain_generalization}\red{c}).

To address both annotation cost and domain generalization issues, we rethink the potential of coarse segmentation data in human matting.
Segmentation labeling is less expensive than matte labeling, and there is a large amount of publicly available segmentation labels for natural images. 
In this paper, we introduce a new learning paradigm for cost-efficient trimap-free human matting, called weakly semi-supervised human matting (WSSHM), to reduce annotation cost using a small number of expensive matte labels and a large number of economical segmentation labels, while improving generalization performance to natural images.

\subsection{Notation}
We denote a segmentation dataset as $\mathcal{D}_{s}$ that consists of a set of pairs $(I^{s}, S)$ where $I^{s} \in \mathbb{R}^{H \times W \times 3}$ is an natural image and $S \in \{0, 1\}^{H \times W}$ is a coarse segmentation label and a matting dataset as $\mathcal{D}_{m}$ that consists of a set of pairs $(I^{m}, M)$ where $I^{m} \in \mathbb{R}^{H \times W \times 3}$ is a synthetic image and $M \in [0, 1]^{H \times W}$ is a fine matte label.
The images in $\mathcal{D}_{s}$ and $\mathcal{D}_{m}$ are totally disjoint.
Here, the $I^{m}$ is composited with a foreground image $I^{fg}$ and a background image $I^{bg}$ as follows:
\begin{equation}
    I^{m} = M \cdot I^{fg}  + (1 - M) \cdot I^{bg}.
\end{equation}
We assume that matting datasets consisting of natural images are unavailable for training, as is the case in real-world scenarios.

\subsection{Baseline Training Method}
We introduce a simple baseline method to achieve the goal of the WSSHM.
As illustrated in \figurename~\ref{fig:overview}, we employ a two-step pipeline involving teacher-student networks~\cite{tarvainen2017mean,berthelot2019mixmatch,(simclr)chen2020simple} to effectively blend the boundary detail representation from matte data with the domain generalization capability from segmentation data.

\paragraph{\textbf{Step 1: Teacher Network Training}}
In the first step, we train the teacher network $f^{t}(\cdot)$ using the synthetic matting dataset $\mathcal{D}_{m}$ to develop robust boundary detail representation. The teacher network generates two outputs: the output matte $M' \in [0, 1]^{H \times W}$ and the output boundary $\alpha' \in \{0, 1\}^{H \times W}$. 
The boundary label indicates regions that require detailed representation, and the ground-truth boundary label $\alpha$ is obtained from the matte label $M$ in $\mathcal{D}_{m}$:
\[ \alpha_{i,j} = \begin{cases} 1 & \text{if } 0.05 < M_{i,j} < 0.95 \\ 0 & \text{otherwise.} \end{cases} \]
We train the network leveraging a conventional matte loss function:
\begin{gather}
    L = L_{matte} + \lambda L_{boundary} \label{eq:lambda}\\
    L_{matte} = L_{mse} + L_{grad} \label{eq:mat_loss}\\
    L_{mse} = (M - M')^2 \\
    L_{grad} = |\nabla_{x}(M) - \nabla_{x}(M')| + |\nabla_{y}(M) - \nabla_{y}(M')| \\
    L_{boundary} = |\alpha-\alpha'|,
\end{gather}
where $\lambda$ is set to 0.01 by default.

\begin{figure*}[t]
    \centering
    \includegraphics[width=0.75\linewidth]{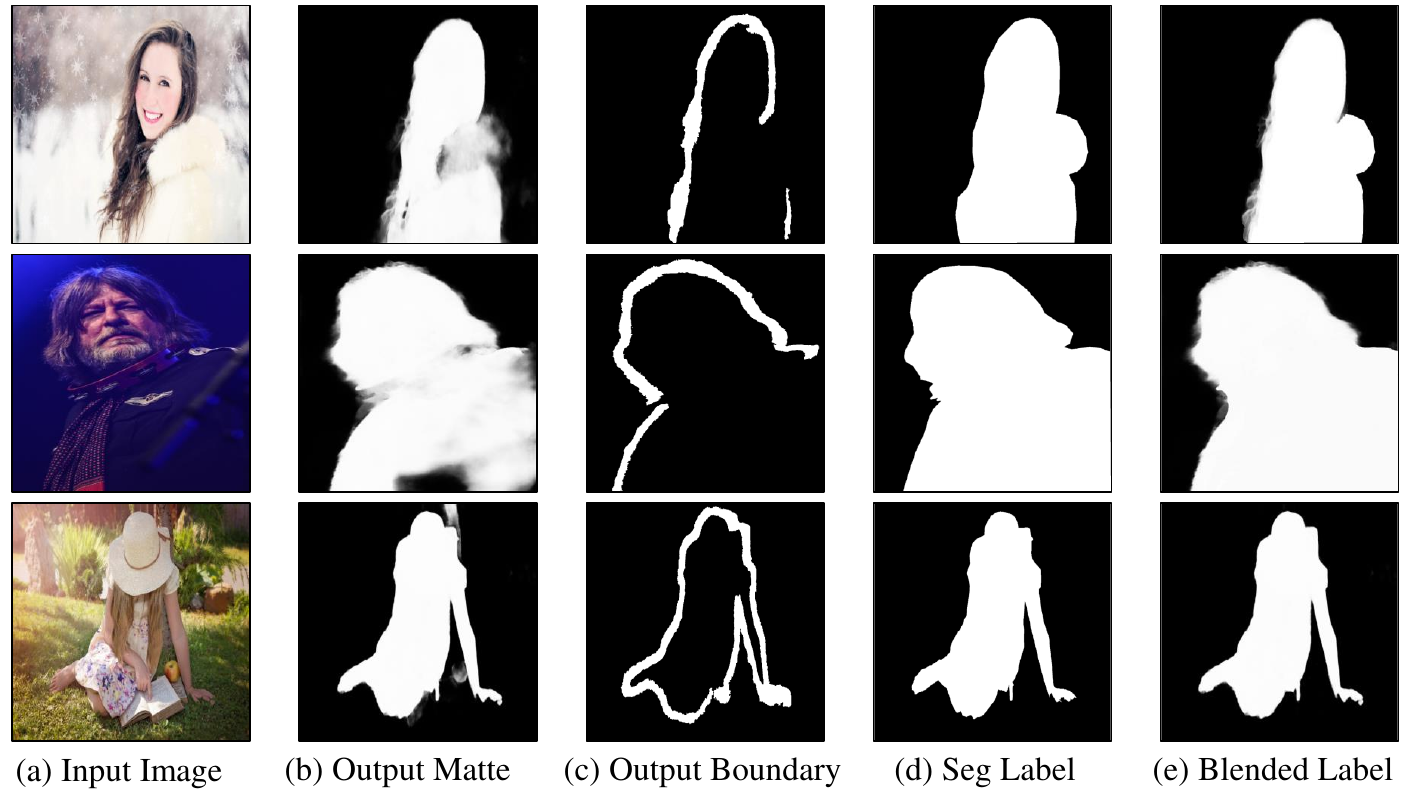}
    \caption{\textbf{Qualitative samples of Matte Label Blending} demonstrating its ability to generate high-quality blended matte labels by effectively combining boundary detail representation from teacher network outputs with the coarse-level knowledge from ground-truth segmentation labels.}
    \label{fig:sample_mlb}
    \vspace{-2mm}
\end{figure*}

\paragraph{\textbf{Step 2: Student Network Training with Matte Label Blending}}
In the second step, we freeze the teacher network and train the student network $f^{s}(\cdot)$ using the natural segmentation dataset $\mathcal{D}_{s}$ to mitigate the domain generalization problem. 
The frozen teacher network provides pseudo matte $M'$ and boundary $\alpha'$ labels for the weakly augmented natural images. 
However, the teacher network often struggles to generalize to natural images due to the synthetic nature of the training data, as shown in \figurename~\ref{fig:sample_domain_generalization}\red{c} and \figurename~\ref{fig:sample_mlb}\red{b}.
Notably, we observe that errors primarily occur in the foreground and background regions, while predictions on the boundary regions are relatively accurate.
Based on this observation, we introduce an effective label blending mechanism, named \textbf{Matte Label Blending (MLB)}.
This mechanism adaptively combines the boundary representation from $M'$ with the coarse-level knowledge from segmentation label $S$, generating a blended matte label $\bar{M}=\alpha'M'+(1-\alpha')S$.
Namely, as shown in \figurename~\ref{fig:sample_mlb}\red{e}, it removes noisy errors in the non-boundary region of $M'$ using segmentation label $S$ while maintaining detailed representation in the boundary region of $M'$, resulting in high-quality pseudo matte labels.

The student network takes strongly augmented natural images and is trained using the blended matte label $\bar{M}$ with the $L_{matte}$ loss function, enabling it to develop robustness to natural images and precise boundary detail representation.
Additionally, we employ exponential moving average (EMA) updates to slowly update the teacher network parameters using those of the student network if both network architectures are the same.
This approach effectively combines the strengths of synthetic matte data and natural segmentation data, providing a comprehensive solution for the WSSHM that improves both boundary detail representation and domain generalization.

\section{Experiments}

\subsection{Datasets and Evaluation Metrics}

\paragraph{\textbf{Training Dataset.}}
For the human matting dataset, we use Human-2K dataset~\cite{(human2k)liu2021tripartite}, which contains 2,000 images and is publicly available.
Note that images in the Human-2K dataset have only foreground region information.
Following the common practice~\cite{(PPM)ke2022modnet, (matteformer)park2022matteformer,(distinctions)(HAttMatting)qiao2020attention, (RWP)(MGMatte)yu2021mask, (LFM)zhang2019late}, we generate the synthesized images by compositing the foreground images of Human-2K dataset with background images of the COCO dataset~\cite{(coco)lin2014microsoft}, which contain no humans to prevent their negative influence on the stable training of trimap-free matting models.
For the human segmentation dataset, we assemble public portrait segmentation datasets~\cite{(supervisely),(tiktok)jafarian2021learning, (eg1800)shen2016automatic, (baidu)wu2014early}. 
Note that all images in the segmentation datasets are natural images.
We set the maximum number of segmentation labels to 50K, and the rest of the samples, except the public datasets, are collected from our private segmentation dataset. 

\paragraph{\textbf{Validation Dataset.}}
To validate the human matting performance, we use three matting datasets which consist of natural images (P3M~\cite{(P3M)li2021privacy}, PPM~\cite{(PPM)ke2022modnet}, and RWP~\cite{(RWP)(MGMatte)yu2021mask}) and one synthetic dataset (D-646~\cite{(distinctions)(HAttMatting)qiao2020attention}).
The P3M, PPM, and RWP consist of 500, 100, and 636 natural scene images.
The D-646 dataset includes 362 foreground images, and we composed these foreground images with five VOC~\cite{(voc)everingham2010pascal} background images.

\paragraph{\textbf{Evaluation Metrics.}}
We adopt the mean square error (MSE) and the sum of absolute differences (SAD) metrics, which are widely used for the human matting task.
For the detailed analysis, we separately measure the scores on the whole and boundary regions.
The pixels that are larger than 0.05 and smaller than 0.95 in the ground-truth matte label are treated as the boundary region pixels, following \cite{(RWP)(MGMatte)yu2021mask}.
We mainly use the MSE metric for analysis and provide the SAD scores in the appendix.

\begin{figure*}[t]
    \begin{minipage}[b]{0.49\linewidth}
        \centering
        \begin{subfigure}[b]{0.49\linewidth}
            \centering
            \includegraphics[width=\linewidth]{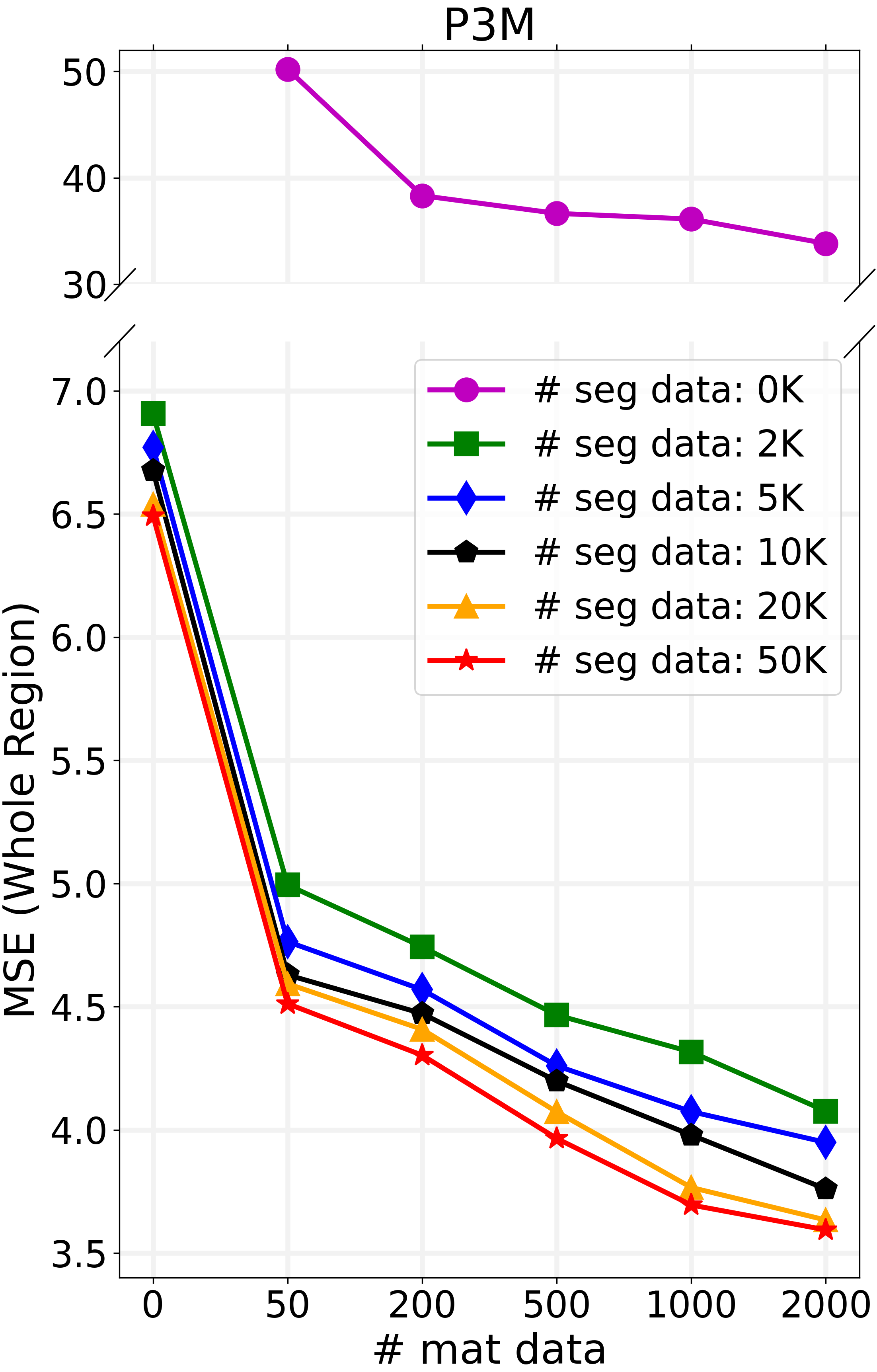} 
            \label{fig:P3M_mse_whole}  
        \end{subfigure}
        \begin{subfigure}[b]{0.49\linewidth}
            \centering
            \includegraphics[width=\linewidth]{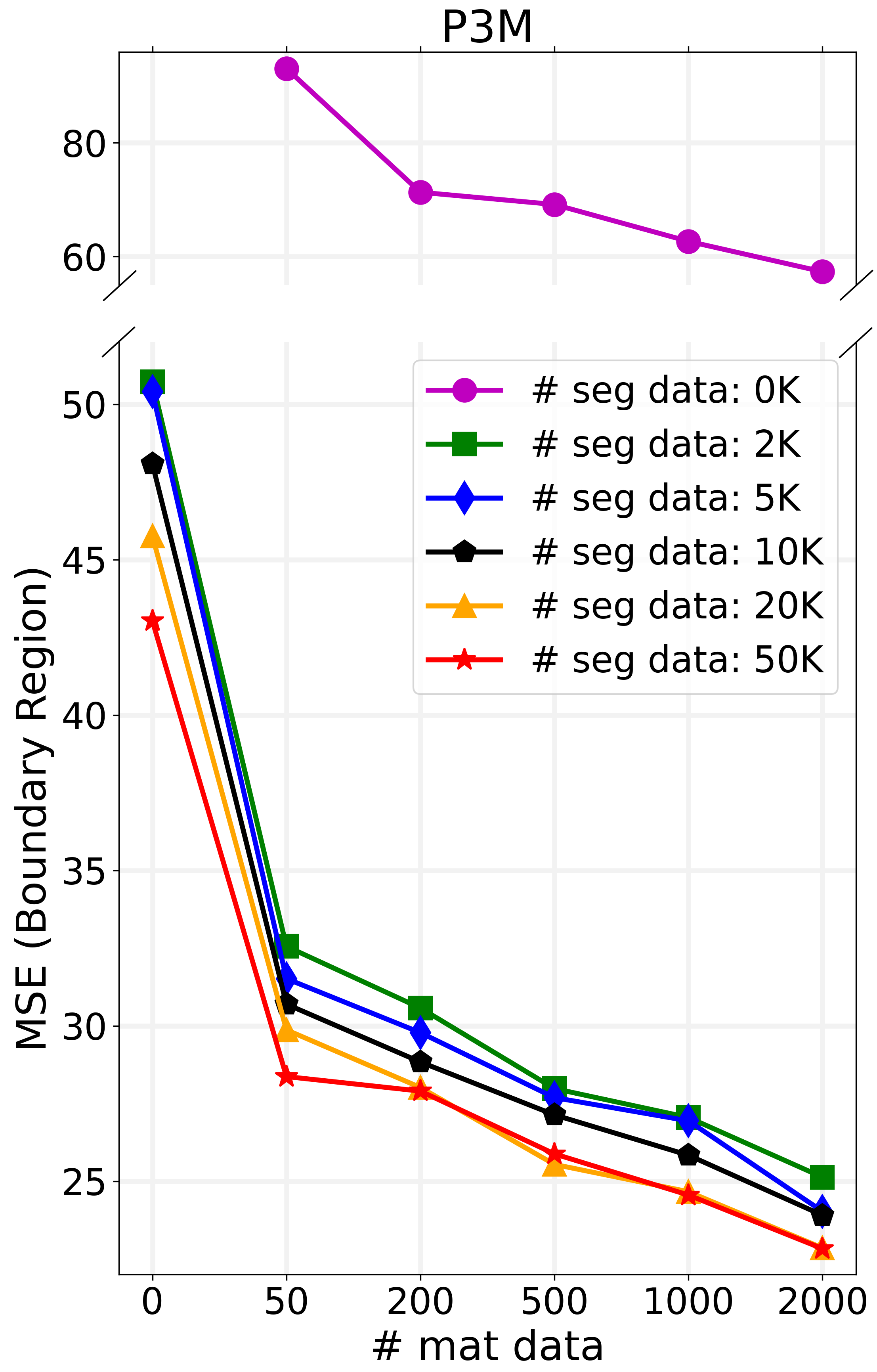} 
            \label{fig:P3M_mse_boundary}  
        \end{subfigure}
        \vspace{-5mm}
        \caption{Comparisons on \textbf{P3M} dataset (\textbf{natural images}) according to the amount of segmentation and matte data.}
        \label{fig:P3M}
        \vspace{4mm}
    \end{minipage}
    \hspace{2mm}
    \begin{minipage}[b]{0.49\linewidth}
        \centering
        \begin{subfigure}[b]{0.49\linewidth}
            \centering
            \includegraphics[width=\linewidth]{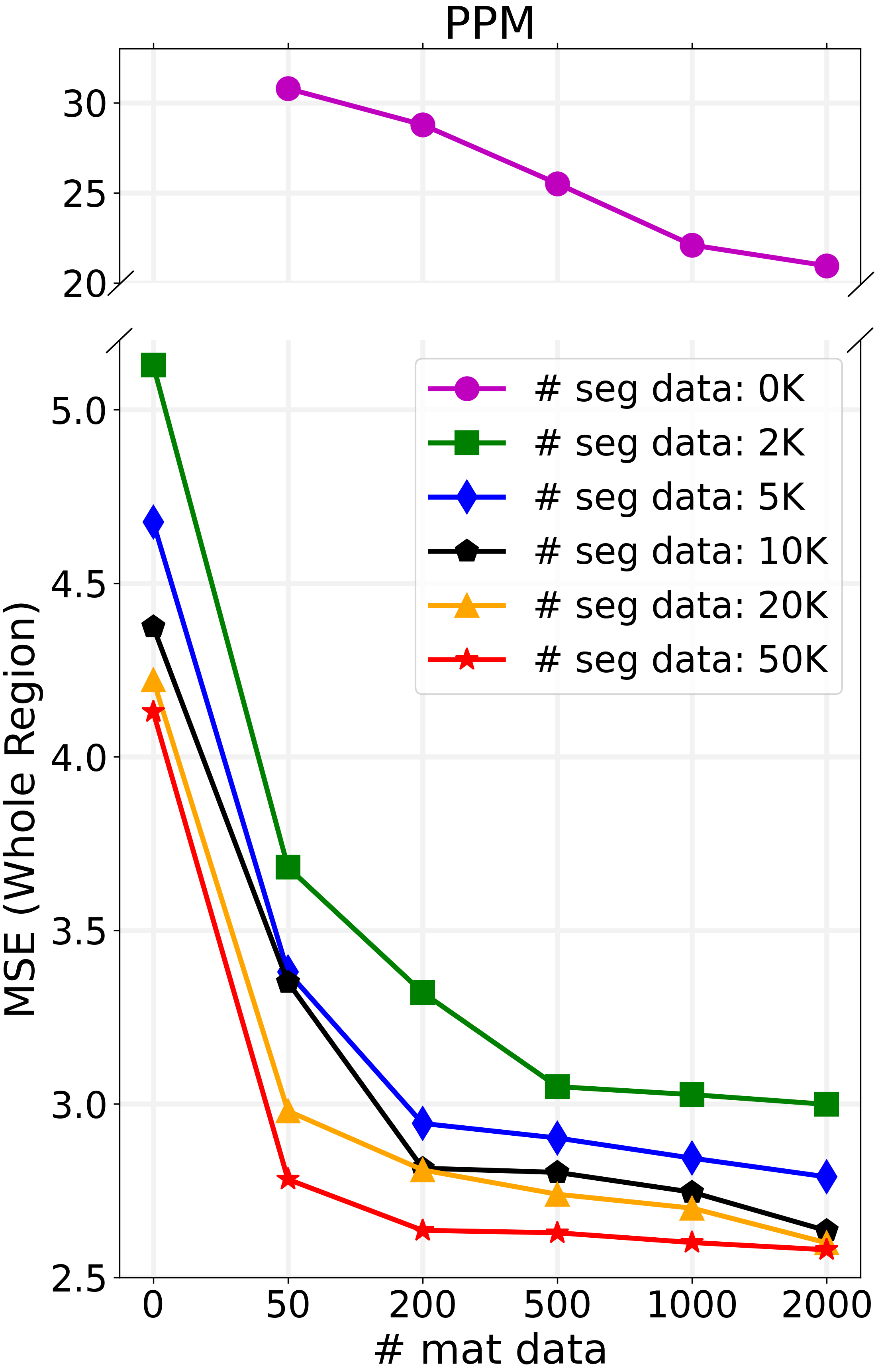} 
            \label{fig:PPM_mse_whole}  
        \end{subfigure}
        \begin{subfigure}[b]{0.49\linewidth}
            \centering
            \includegraphics[width=\linewidth]{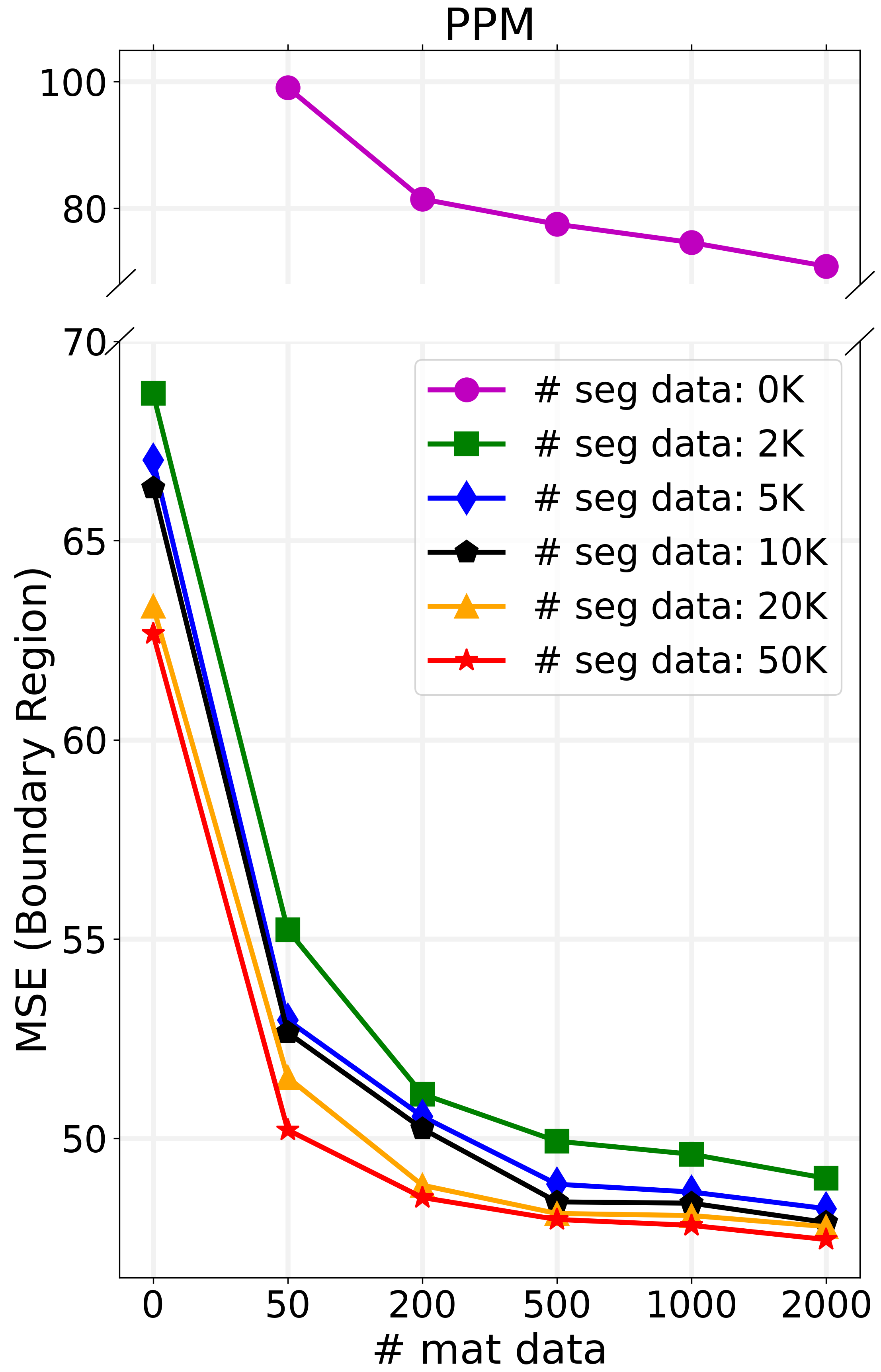} 
            \label{fig:PPM_mse_boundary}  
        \end{subfigure}
        \vspace{-5mm}
        \caption{Comparisons on \textbf{PPM} dataset (\textbf{natural images}) according to the amount of segmentation and matte data.}
        \label{fig:PPM}
        \vspace{4mm}
    \end{minipage}
    \hspace{2mm}
    \begin{minipage}[b]{0.49\linewidth}
        \centering
        \begin{subfigure}[b]{0.49\linewidth}
            \centering
            \includegraphics[width=\linewidth]{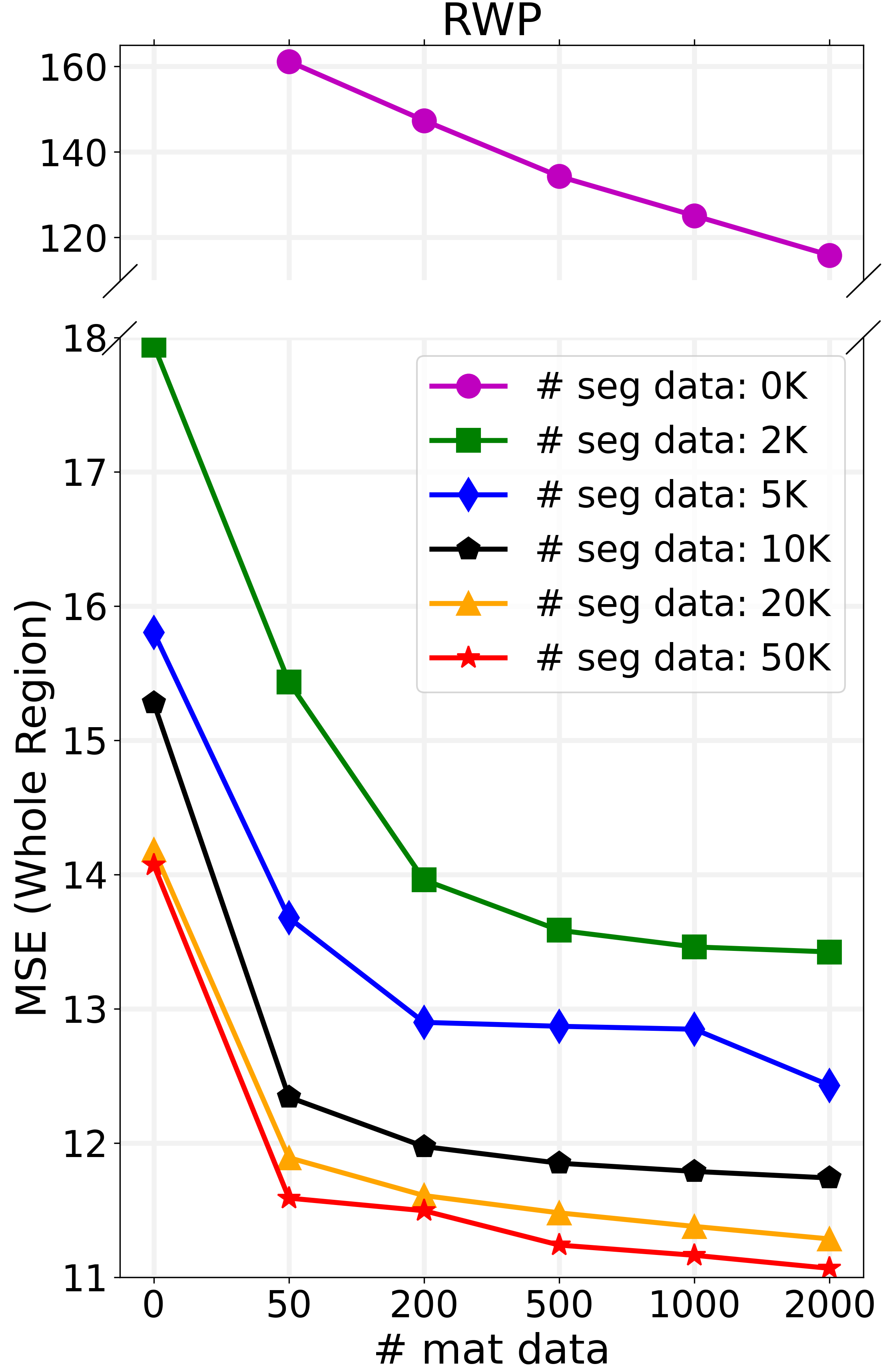} 
            \label{fig:RWP_mse_whole}  
        \end{subfigure}
        \begin{subfigure}[b]{0.49\linewidth}
            \centering
            \includegraphics[width=\linewidth]{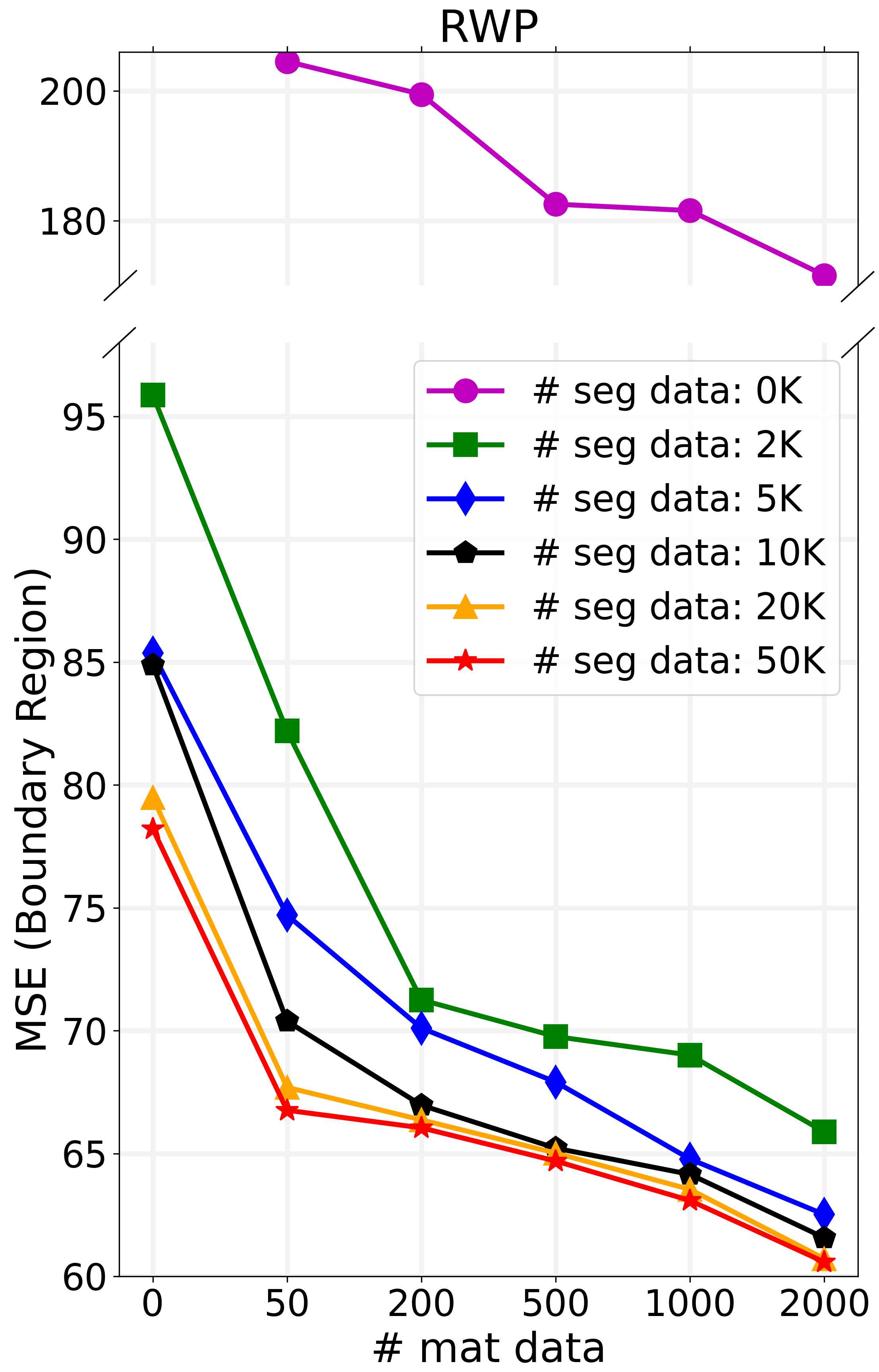} 
            \label{fig:RWP_mse_boundary}  
        \end{subfigure}
        \vspace{-5mm}
        \caption{Comparisons on \textbf{RWP} dataset (\textbf{natural images}) according to the amount of segmentation and matte data.}
        \label{fig:RWP}
        \vspace{-2mm}
    \end{minipage}
    \hspace{2mm}
    \begin{minipage}[b]{0.49\linewidth}
        \centering
        \begin{subfigure}[b]{0.49\linewidth}
            \centering
            \includegraphics[width=\linewidth]{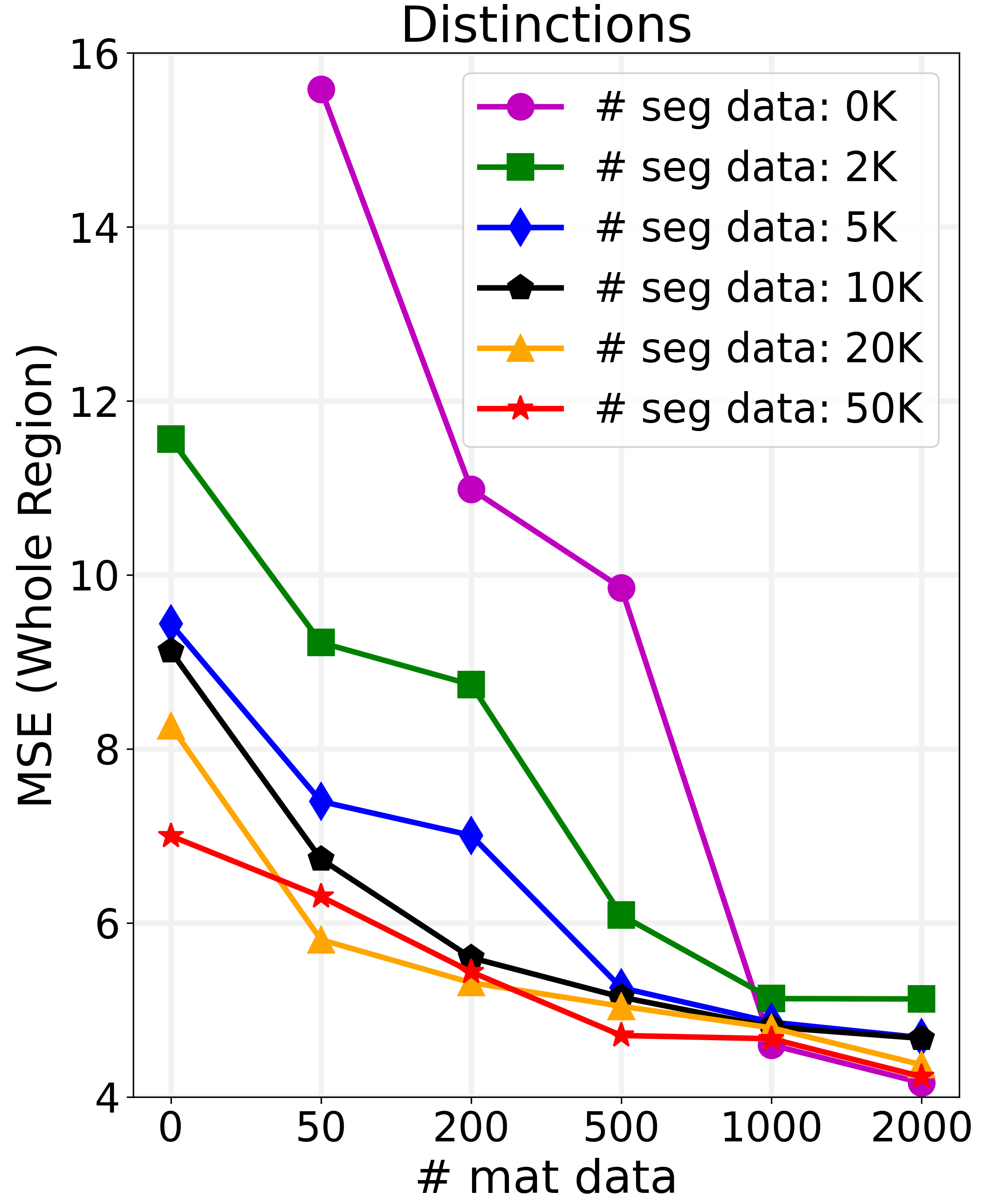} 
            \label{fig:D646_mse_whole}  
        \end{subfigure}
        \begin{subfigure}[b]{0.49\linewidth}
            \centering
            \includegraphics[width=\linewidth]{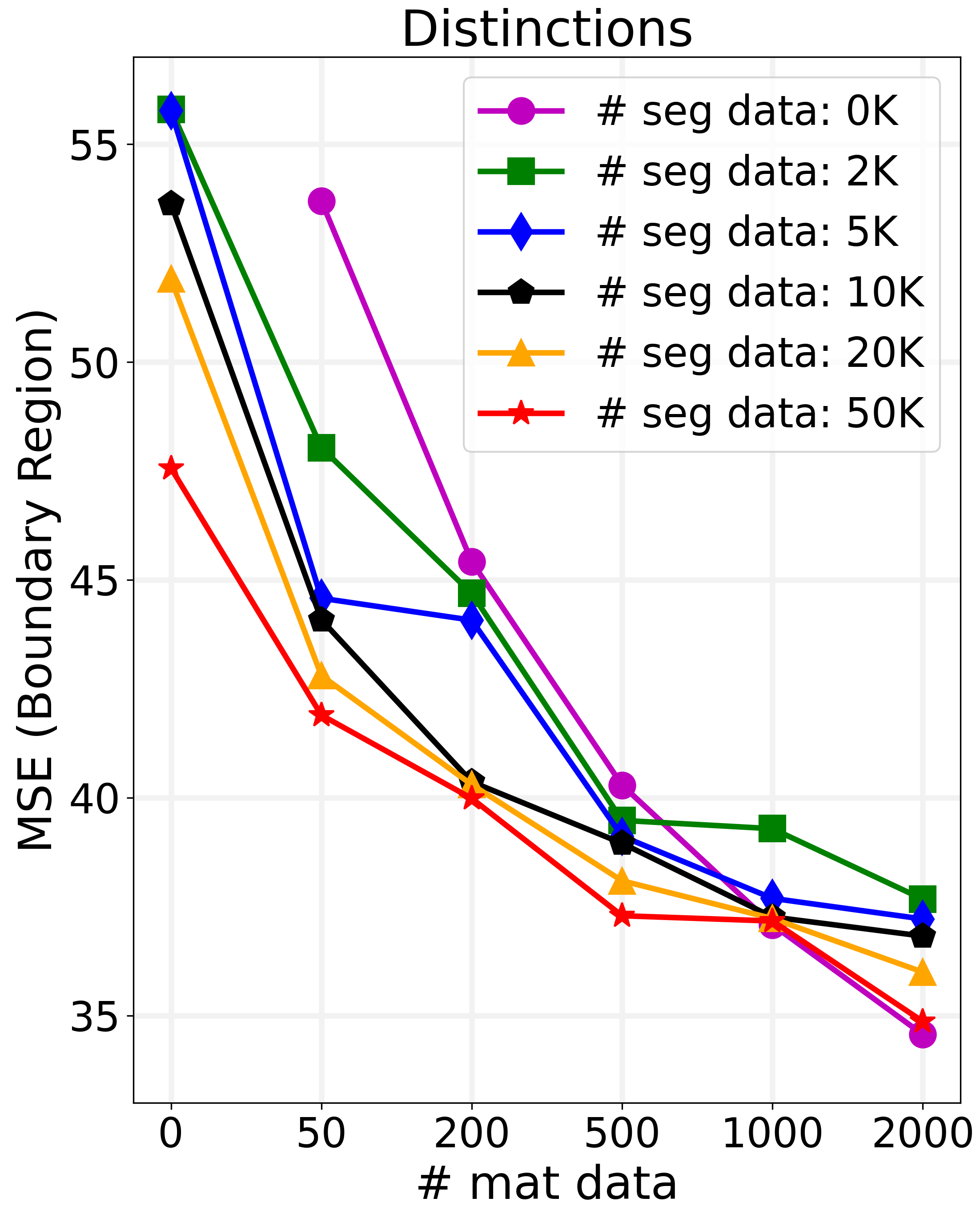} 
            \label{fig:D646_mse_boundary}  
        \end{subfigure}
        \vspace{-5mm}
        \caption{Comparisons on \textbf{D-646} dataset (\textbf{synthetic images}) according to the amount of segmentation and matte data.}
        \label{fig:D646}
        \vspace{-2mm}
    \end{minipage}
\end{figure*}

\subsection{Implementation Details}

\paragraph{\textbf{Baseline Network.}}
We employ the U-Net~\cite{(unet)ronneberger2015u} architecture with ResNet~\cite{(resnet)he2016deep} backbone and an Atrous Spatial Pyramid Pooling~\cite{(aspp)chen2017deeplab} module as a baseline matting network, which is commonly used in existing matting approaches~\cite{(shm)chen2018semantic,(human2k)liu2021tripartite,(RWP)(MGMatte)yu2021mask,(LFM)zhang2019late} and quite straightforward architecture.
We use the U-Net with ResNet-101 backbone for the teacher and student networks by default.
The student network can be flexibly replaced with other networks.
To verify the applicability of our training method on real-time models, we also adopt the ResNet-18 or ResNet-18 with a width multiplier of 0.5 backbones for the student network.

\paragraph{\textbf{Optimization.}}
We use the Adam optimizer, cosine learning rate decay scheduling, and a batch size of 16 to train the network.
We initialize the parameters of both teacher and student networks with the pre-trained weights, which are trained on segmentation datasets, for fast and stable convergence.
Note that the pre-trained weights are not used when training the model using only the matting dataset.
During the pre-training stage, we set the learning rate to 1e-4 and training iterations to 200K using the MSE objective function.
The teacher network is fine-tuned on the synthetic matting dataset, the learning rate is set to 5e-5, and the training iteration is 10K with the objective function of Eq. \ref{eq:mat_loss}.
After training the teacher network, we train the student network using the natural segmentation dataset with the Matte Label Blending strategy.
Here, we set the learning rate to 5e-5 and training iterations to 20K.
We employ random scaling, random cropping, and random horizontal flipping for basic data augmentation and apply the color jittering operations~\cite{(simclr)chen2020simple} only for strong augmentation.
We also adapt the multi-scale training strategy by cropping the image from $512\times512$ to $768\times768$.
Note that our final result is generated only from the student network.
During the test time, we resize the image to the shorter edge length of the image to be 512.
We train the models on 4 NVIDIA V100 GPUs using Pytorch~\cite{(pytorch)paszke2019pytorch} framework.

\begin{table*}[t]
  \centering
  \begin{adjustbox}{max width=0.69\linewidth}
  \begin{tabular}{c|c|c|c|c|c|c}
    \toprule
    \multirow{2}{*}{Method} & \multicolumn{4}{c|}{Whole Region MSE $\downarrow$} & \multicolumn{2}{c}{FPS $\uparrow$} \\
    \cline{2-7} \addlinespace[0.5ex]
                            & P3M~\cite{(P3M)li2021privacy} & PPM~\cite{(PPM)ke2022modnet} & RWP~\cite{(RWP)(MGMatte)yu2021mask} & D-646~\cite{(distinctions)(HAttMatting)qiao2020attention} & GPU & CPU \\
    \midrule
    SHM~\cite{(shm)chen2018semantic} & 40.966 & 27.306 & 58.456 & 23.646 & 97.0 & 6.6\\
    GCA~\cite{li2020natural} & 22.176 & 16.337 & 32.423 & 11.523 & 95.1 & 5.0 \\
    P3MNet~\cite{(P3M)li2021privacy} & 13.586 & 9.672 & 24.079 & 8.935 & 94.6 & 11.8 \\
    IndexNet~\cite{lu2019indices} & 12.484 & 6.043 & 20.580 & 7.683 & 52.9 & 1.9 \\
    GFM~\cite{(gfm)li2022bridging} & 10.770 & 5.183 & 18.895 & 6.705 & 100.3 & 13.3 \\
    LFM~\cite{(LFM)zhang2019late} & 9.611 & 5.970 & 18.012 & 6.222 & 26.3 & 2.0 \\
    MODNet~\cite{(PPM)ke2022modnet} & 9.692 & 5.161 & 18.038 & \textbf{\blue{5.736}} & 101.7 & 8.9 \\
    \midrule
    \textbf{Ours (R101)} & \textbf{\red{4.391}} & \textbf{\red{2.802}} & \textbf{\red{11.257}} & \textbf{\red{5.301}} & 58.3 & 5.0 \\
    \textbf{Ours (R18)} & \textbf{\blue{7.007}} & \textbf{\blue{3.518}} & \textbf{\blue{13.094}} & 6.192 & \textbf{\blue{222.2}} & \textbf{\blue{18.7}} \\
    \textbf{Ours (R18 0.5)} & 8.307 & 3.996 & 16.787 & 7.816 & \textbf{\red{328.9}} & \textbf{\red{43.7}} \\
    \bottomrule
  \end{tabular}
  \end{adjustbox}
  \caption{
    \textbf{Comparison with state-of-the-art trimap-free matting approaches.}
    All models are trained with the same amount of 10K segmentation and 200 matte labels.
    The P3M~\cite{(P3M)li2021privacy}, PPM~\cite{(PPM)ke2022modnet}, and RWP~\cite{(RWP)(MGMatte)yu2021mask} datasets consist of natural images, and the D-646~\cite{(distinctions)(HAttMatting)qiao2020attention} consists of synthetic images.
    The R18 (0.5) denotes UNet ResNet18 backbone network with a width multiplier of 0.5.
  }
  \label{tab:comparison_matting_methods}
\end{table*}
\begin{figure}[t]
    \centering
    \includegraphics[width=\linewidth]{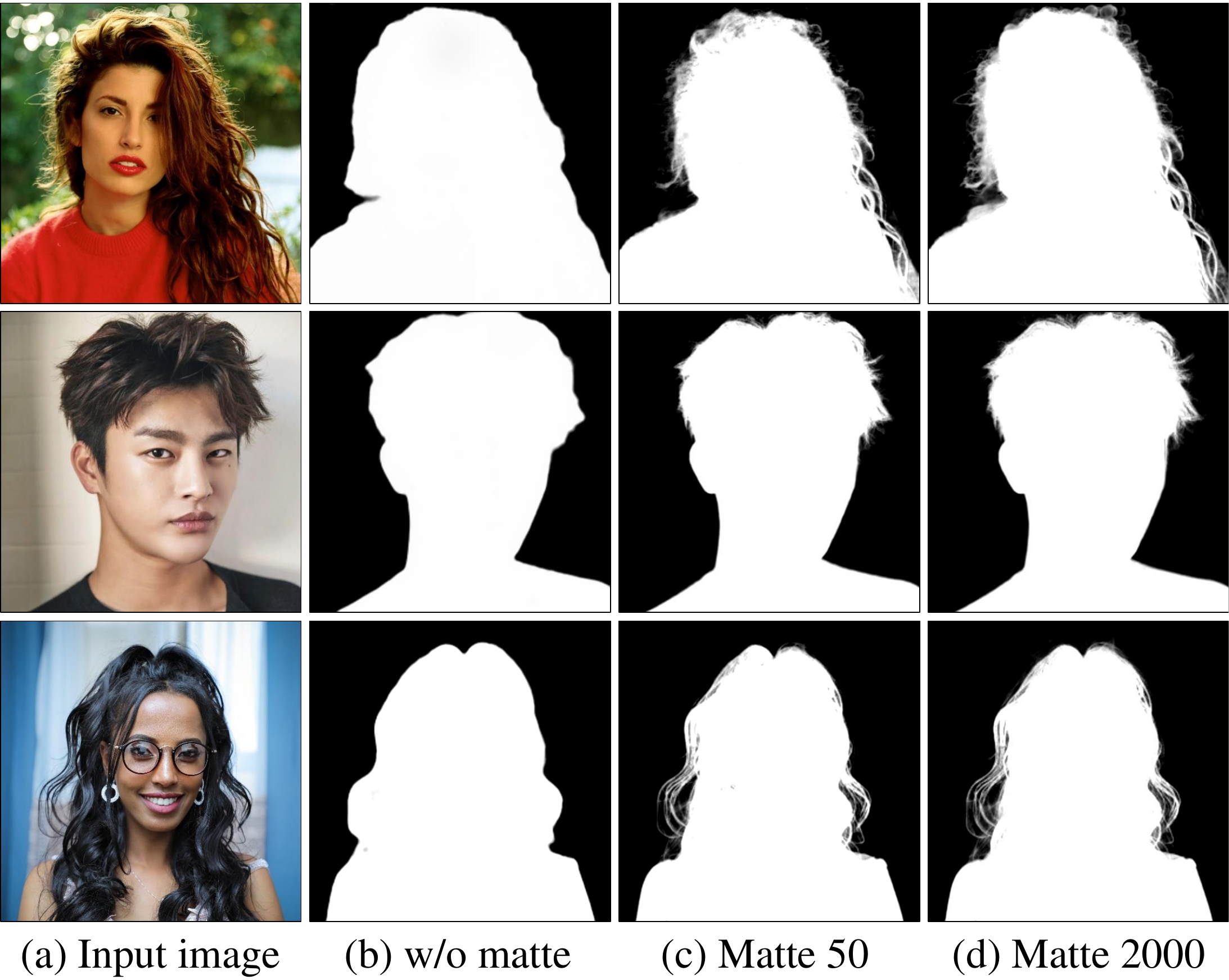}
    \caption{\textbf{Qualitative comparisons} according to the amount of matte data with a fixed number of 10K segmentation data: (b) without matte data, (c) 50 matte data, (d) 2000 matte data.}
    \label{fig:effect_num_matte_labels}
    \vspace{-2mm}
\end{figure}

\subsection{Results}
We conducted extensive experiments on the proposed method to analyze the effect of varying amounts of matte and segmentation labels. 
We split segmentation data into six subsets with 0, 2K, 5K, 10K, 20K, and 50K labels, and matte data into six subsets with 0, 50, 200, 500, 1000, and 2000 labels. 
We train the model on all ({=}6{$\times$}6{-}1) combinations of the segmentation and matte subsets, and evaluate on four validation sets, namely P3M, PPM, RWP, and D-646. 
We primarily analyze the performance of the model on natural images using P3M (\figurename~\ref{fig:P3M}), PPM (\figurename~\ref{fig:PPM}), and RWP (\figurename~\ref{fig:RWP}) datasets and on synthetic images using D-646 dataset (\figurename~\ref{fig:D646}). 
Note that there is no pre-training for segmentation subsets with zero data counts.

\paragraph{\textbf{Finding 1: Segmentation data improves the robustness of the matting model to natural images.}}
The model trained with only matte data ($e.g.,$ seg 0 \& mat 2000) achieves a reasonable performance on synthetic images, as shown in \figurename~\ref{fig:sample_domain_generalization}\red{c} and \figurename~\ref{fig:D646}.
However, it shows completely different results on natural images, as shown in \figurename~\ref{fig:sample_domain_generalization}\red{c} and Figures~\ref{fig:P3M}\red{--}\ref{fig:RWP}.
The reason is that the model often fails to generalize to natural images because it is trained with only synthetic matte data.
When we leverage segmentation data with the proposed training method ($e.g.,$ seg 2000 \& mat 2000), the performance of the model dramatically improves from 34.2 MSE to 4.1 MSE.
We also discover that the performance gap between using 2K and 50K segmentation data is not substantial (4.1 vs. 3.6 MSE).
The results demonstrate that \textbf{(1)} domain generalization is a critical issue for trimap-free matting models when training with only synthetic matte data, and \textbf{(2)} although segmentation data consists of weaker information, our method can leverage segmentation data as an effective source to improve the robustness of the model to natural images.

\paragraph{\textbf{Finding 2: A small amount of matte data can dramatically improve the boundary detail representation of the matting model.}}
The model trained with only segmentation data ($e.g.,$ seg 10K \& mat 0) performs well on natural images but lacks representing boundary detail information due to the coarse training labels, as shown in \figurename~\ref{fig:effect_num_matte_labels}\red{b}.
Even when we increase the amount of segmentation data, the improvement of the model in the boundary region is marginal.
However, by leveraging a small amount of synthetic matte data, our training method considerably improves the prediction accuracy of the model, especially on the boundary region.
As shown in the results on P3M dataset consisting of natural images (\figurename~\ref{fig:P3M}), the MSE error on the boundary region is steeply decreased when the amount of matte data is increased from 0 to 50.
Moreover, we found that the performance gap between leveraging 50 and 2000 matte data is insignificant, as shown in \figurename~\ref{fig:effect_num_matte_labels}\red{c} and \figurename~\ref{fig:effect_num_matte_labels}\red{d}.
The results demonstrate that \textbf{(1)} incorporating a small amount of matte data using our method can dramatically improve the boundary detail representation of the model, and \textbf{(2)} it gives us more opportunity for cost-efficient human matting using a reduced amount of segmentation and matte data.

\paragraph{\textbf{Comparison with state-of-the-art methods.}}
Our method is compared to state-of-the-art trimap-free matting methods such as SHM~\cite{(shm)chen2018semantic}, GCA~\cite{li2020natural}, GFM~\cite{(gfm)li2022bridging}, P3MNet~\cite{(P3M)li2021privacy}, IndexNet~\cite{lu2019indices}, LFM~\cite{(LFM)zhang2019late}, and MODNet~\cite{(PPM)ke2022modnet} as in Table \ref{tab:comparison_matting_methods} using the same amount of 10K segmentation and 200 matte labels.
Following the training strategy used in \cite{(PPM)ke2022modnet}, all existing methods are pre-trained on segmentation labels and fine-tuned on matte labels.
Our three baseline networks are trained using the proposed training method, Matte Label Blending.
In addition, we measure the throughput of each model using ONNX runtime~\cite{onnxruntime} on an NVIDIA Tesla V100 GPU and Xeon Gold 5120 CPU with an input size of 512{$\times$}512.
We evaluate all methods using three real-world datasets ($i.e.,$ P3M, PPM, and RWP) consisting of natural images and one synthesized dataset ($ i.e.,$ D-646) consisting of synthetic images.
As a result, the performance of ours on the synthesized dataset is not substantial compared to state-of-the-art matting methods.
However, we achieve remarkable performances on real-world datasets since our training method can noticeably improve the domain generalization of the model to natural images.
Especially, our training method is successfully applied to the lightweight model ($i.e.,$ UNet-ResNet18 with a multiplier of 0.5) achieving extremely fast inference of 328.9 FPS on a GPU and state-of-the-art performance.

\begin{table}[t]
      \centering
      \begin{adjustbox}{max width=0.95\linewidth}
      \begin{tabular}{c|c|c}
        \toprule
        \multirow{2}{*}{Network} & Whole Region & Boundary Region \\
        \cline{2-3} \addlinespace[0.5ex]
                & MSE $\downarrow$ & MSE $\downarrow$  \\
        \midrule
        \multicolumn{3}{c}{\textit{\textbf{seg 2K, mat 50}}} \\
        BSHM~\cite{(boosting)liu2020boosting} & 20.012 & 87.186 \\
        Context-aware~\cite{hou2019context} & 19.837 & 52.943 \\
        \textbf{ours}  & \textbf{4.901} & \textbf{32.435}  \\
        \midrule
        \multicolumn{3}{c}{\textit{\textbf{seg 10K, mat 200}}} \\
        BSHM~\cite{(boosting)liu2020boosting} & 15.779 & 80.787 \\
        Context-aware~\cite{hou2019context} & 16.844 & 46.361 \\
        \textbf{ours}  & \textbf{4.391} & \textbf{28.776}  \\
        \midrule
        \multicolumn{3}{c}{\textit{\textbf{seg 50K, mat 2000}}} \\
        BSHM~\cite{(boosting)liu2020boosting} & 13.287 & 75.600 \\
        Context-aware~\cite{hou2019context} & 14.092 & 42.006 \\
        \textbf{ours}  & \textbf{3.512} & \textbf{22.728}  \\
        \bottomrule
      \end{tabular}
      \end{adjustbox}
      \caption{
        \textbf{Comparison with existing training methods}, BSHM~\cite{(boosting)liu2020boosting} and Context-aware~\cite{hou2019context}, for human matting using the same amount of segmentation and matte data.
      }
      \label{tab:comparison_training_methods}
\end{table}
Also, our training method is compared to existing training methods such as BSHM~\cite{(boosting)liu2020boosting} and Context-aware~\cite{hou2019context} that are proposed to mitigate the domain generalization issue.
We reproduce them using the same amount of segmentation and matte data and evaluate them on P3M validation set.
The results in Table~\ref{tab:comparison_training_methods} demonstrate the substantial superiority of our training method.
The reason is that when BSHM sequentially trains three sub-network in an offline manner, the last sub-network is trained with only synthetic matte data, resulting in inevitable domain generalization errors.
Also, Context-aware consists of simple image augmentation methods, such as re-JPEGing and Gaussian blur. Such image augmentation may remove some artifacts in synthesized images, but cannot handle the contextual unnaturalness as shown in \figurename~\ref{fig:form_seg_and_matte_labels} and \figurename~\ref{fig:sample_domain_generalization}.

\subsection{Analysis}
We conduct experiments to analyze each component of our training method on P3M~\cite{(P3M)li2021privacy} validation set under the segmentation 10K and matte 200 labels setup.

\begin{figure}[t]
    \centering
    \includegraphics[width=\linewidth]{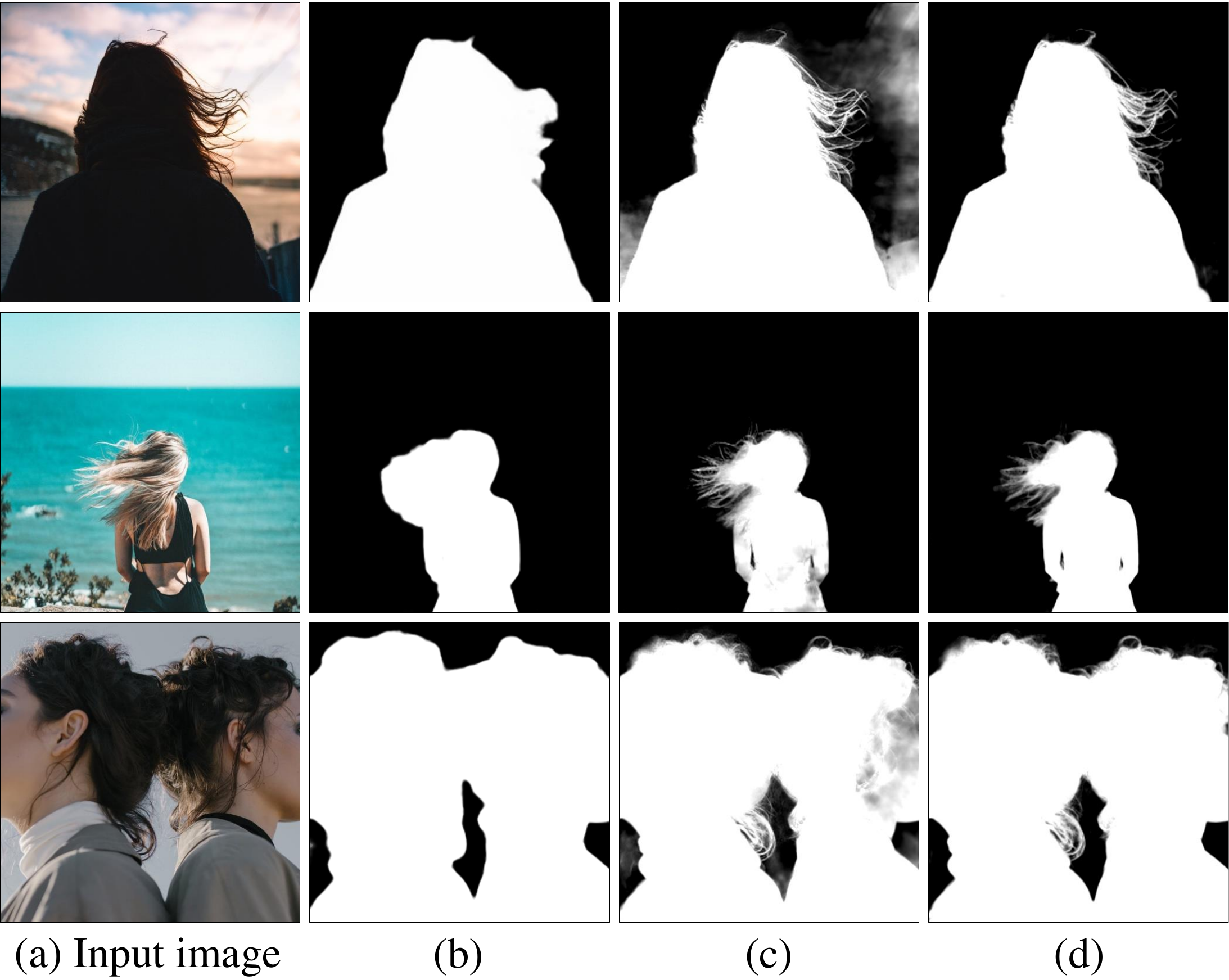}
    \caption{\textbf{Qualitative comparisons} under 50K segmentation data and 2K matte data setting: (b) teacher network trained with only segmentation data, (c) teacher network fine-tuned on matte data, (d) student network trained with Matte Label Blending. }
    \label{fig:sample_networks}
    \vspace{5mm}
\end{figure}

\paragraph{\textbf{Effect of Matte Label Blending.}} 
We compare the performance of the student network with that of the teacher network to investigate the effect of the proposed method.
Table \ref{tab:teacher_student} and \figurename~\ref{fig:sample_networks} show the comparison results between the student network and two kinds of teacher networks.
The first is the teacher network trained with only segmentation data, and the second is the teacher network that is fine-tuned on the synthetic matte data.
The teacher network trained with only segmentation data shows a reasonable performance on the whole region, but the boundary detail representation is insufficient, as in Table \ref{tab:teacher_student} and \figurename~\ref{fig:sample_networks}\red{b}.
When we fine-tune the teacher network using matte data, the boundary detail representation of the network is improved, but there are several noisy predictions in the foreground and background region, dropping the performance in the whole region, as in Table \ref{tab:teacher_student} and \figurename~\ref{fig:sample_networks}\red{c}.
When fine-tuning the network, the continuous guiding of synthetic images deteriorates the generalization of the network to natural images.
However, the Matte Label Blending is designed to exploit only the advantageous information of the teacher network and segmentation data, and the student network can enjoy a substantial improvement on both the whole and boundary region.

\begin{table}[t]
  \centering
  \begin{adjustbox}{max width=\linewidth}
  \begin{tabular}{c|c|c}
    \toprule
    \multirow{2}{*}{Network} & Whole Region & Boundary Region \\
    \cline{2-3} \addlinespace[0.5ex]
            & MSE $\downarrow$ & MSE $\downarrow$  \\
    \midrule
    Teacher (seg only) & 6.676 & 48.088  \\
    Teacher (fine-tune)  & 6.991 & 32.553  \\
    \textbf{Student (ours)}  & \textbf{4.391} & \textbf{28.776}  \\
    \bottomrule
  \end{tabular}
  \end{adjustbox}
  \caption{
    \textbf{Effect of Matte Label Blending.} Note that the Matte Label Blending method is applied only to the student network.
    }
  \label{tab:teacher_student}
\end{table}

\paragraph{\textbf{Effect of loss weight $\lambda$.}}
We conduct experiments to verify the robustness of the model to the hyperparameter $\lambda$ used in Eq. \ref{eq:lambda}.
The experimental results in Table \ref{tab:loss_weight} indicate that our method exhibits consistency across different values of $\lambda$.
Consequently, we set $\lambda$ to 0.01 by default, which yields optimal results across our experiments.

\paragraph{\textbf{Effect of components in the teacher-student framework.}} 
We designed the baseline training method encompassing teacher-student networks, motivated by \cite{tarvainen2017mean,berthelot2019mixmatch,(simclr)chen2020simple}.
Here, we analyze the effect of weak-strong augmentation and exponential moving average (EMA) update strategies used in the teacher-student framework on the performance of human matting models.
The results in Table~\cref{tab:abs_teacher_student} indicate that both strategies yield a slight performance improvement.
We employ both strategies in our training by default and note that the EMA update is applied only when the teacher and student networks have the same network architecture.

\begin{table}[t]
  \centering
  \begin{adjustbox}{max width=0.95\linewidth}
  \begin{tabular}{c|c|c}
    \toprule
    \multirow{2}{*}{$\lambda$} & Whole Region & Boundary Region \\
    \cline{2-3} \addlinespace[0.5ex]
            & MSE $\downarrow$ & MSE $\downarrow$  \\
    \midrule
    0.1 & 4.442 & 28.911  \\
    \textbf{0.01}  & \textbf{4.391} & \textbf{28.776}  \\
    0.001  & 4.421 & 28.838  \\
    \bottomrule
  \end{tabular}
  \end{adjustbox}
  \caption{
    \textbf{Effect of loss weight $\lambda$}. Our training method is robust to changes in $\lambda$, and we set $\lambda$ to 0.01 by default.
  }
  \label{tab:loss_weight}
  \vspace{5mm}
\end{table}

\begin{table}[t]
  \centering
  \begin{adjustbox}{max width=\linewidth}
  \begin{tabular}{c|c|c|c}
    \toprule
    Weak-Strong & EMA & Whole Region & Boundary Region \\
    Aug & Update & MSE $\downarrow$ & MSE $\downarrow$  \\
    \midrule
    \checkmark & \checkmark & \textbf{4.391} & \textbf{28.776}  \\
    \checkmark &  & 4.419 & 28.793  \\
     & \checkmark & 4.437 & 28.804  \\
     &            & 4.472 & 28.815  \\
    \bottomrule
  \end{tabular}
  \end{adjustbox}
  \caption{
    \textbf{Effect of training strategies: weak-strong augmentation and EMA update.} We follow the detailed setting of them used in teacher-student frameworks~\cite{(simclr)chen2020simple,berthelot2019mixmatch,tarvainen2017mean}.
  }
  \label{tab:abs_teacher_student}
\end{table}

\paragraph{\textbf{Transferring to existing matting models.}}
Our training method can be easily transferred to existing matting models by changing the student network in \figurename~\ref{fig:overview}.
We adopt seven existing matting models (SHM~\cite{(shm)chen2018semantic}, MODNet~\cite{(PPM)ke2022modnet}, P3MNet~\cite{(P3M)li2021privacy}, GFM~\cite{(gfm)li2022bridging}, GCA~\cite{li2020natural}, IndexNet~\cite{lu2019indices}, and LFM~\cite{(LFM)zhang2019late}) and compare our training method with the fine-tuning technique, where the model is pre-trained on segmentation data and then fine-tuned on matte data.
As shown in Table~\ref{tab:builtin}, our training method improves the performance of all matting models with a large margin.
This result demonstrates that ours can be served as an effective training method to properly incorporate segmentation and matte data.
Notably, the state-of-the-art real-time matting model, MODNet, shows an improved performance of 5.602 MSE with a 42.2\% reduction in error. 

\begin{table}[t]
  \centering
  \begin{adjustbox}{max width=\linewidth}
  \begin{tabular}{c|c|c|c|c}
    \toprule
    Network & Whole Region & Boundary Region & \multicolumn{2}{c}{FPS} \\
    \cline{2-5} \addlinespace[0.5ex]
            & MSE $\downarrow$ & MSE $\downarrow$ & GPU $\uparrow$ & CPU $\uparrow$ \\
    \midrule
    UNet-R101 & 6.991 & 32.553 & \multirow{2}{*}{58.3} & \multirow{2}{*}{5.0} \\
    + ours & 4.472 \greenpscript{\(-36.0\%\)} & 28.837 \greenpscript{\(-11.4\%\)} &  &  \\
    \midrule
    UNet-R18 & 9.868  & 44.735 & \multirow{2}{*}{222.2} & \multirow{2}{*}{18.7} \\
    + ours & 7.090 \greenpscript{\(-28.2\%\)} & 30.303 \greenpscript{\(-32.3\%\)} &  &  \\
    \midrule
    UNet-R18(0.5) & 11.388 & 55.745 & \multirow{2}{*}{328.9} & \multirow{2}{*}{43.7} \\
    + ours & 8.335 \greenpscript{\(-26.8\%\)} & 31.059 \greenpscript{\(-44.3\%\)} &  &  \\
    \midrule
    MODNet~\cite{(PPM)ke2022modnet} & 9.692 & 42.270 & \multirow{2}{*}{101.7} & \multirow{2}{*}{8.9} \\
    + ours & 5.655 \greenpscript{\(-41.7\%\)} & 34.335 \greenpscript{\(-18.8\%\)} &  &  \\
    \midrule
    LFM~\cite{(LFM)zhang2019late} & 9.611 & 42.737 & \multirow{2}{*}{26.3} & \multirow{2}{*}{2.0} \\
    + ours & 6.103 \greenpscript{\(-36.5\%\)} & 29.633 \greenpscript{\(-30.7\%\)} &  &  \\
    \midrule
    GFM~\cite{(gfm)li2022bridging} & 10.770 & 43.055 & \multirow{2}{*}{100.3} & \multirow{2}{*}{13.3} \\
    + ours & 6.425 \greenpscript{\(-40.3\%\)} & 32.098 \greenpscript{\(-25.4\%\)} &  &  \\
    \midrule
    P3MNet~\cite{(P3M)li2021privacy} & 13.586 & 50.146 & \multirow{2}{*}{94.6} & \multirow{2}{*}{11.8} \\
    + ours & 7.381 \greenpscript{\(-45.7\%\)} & 35.092 \greenpscript{\(-30.0\%\)} &  &  \\
    \midrule
    IndexNet~\cite{lu2019indices} & 12.484 & 46.936 & \multirow{2}{*}{52.9} & \multirow{2}{*}{1.9} \\
    + ours & 7.554 \greenpscript{\(-39.5\%\)} & 34.596 \greenpscript{\(-26.3\%\)} &  &  \\
    \midrule
    GCA~\cite{li2020natural} & 22.176 & 65.232 & \multirow{2}{*}{95.1} & \multirow{2}{*}{5.0} \\
    + ours & 8.875 \greenpscript{\(-60.0\%\)} & 36.712 \greenpscript{\(-43.7\%\)} &  &  \\
    \midrule
    SHM~\cite{(shm)chen2018semantic} & 40.966 & 106.338 & \multirow{2}{*}{97.01} & \multirow{2}{*}{6.6} \\
    + ours & 19.073 \greenpscript{\(-53.4\%\)} & 46.899 \greenpscript{\(-52.0\%\)} &  &  \\
    \bottomrule
  \end{tabular}
  \end{adjustbox}
    \caption{
    \textbf{Transferring our training method to existing matting networks.} \textit{+ours} denotes that the network is trained as the student network in our training method.
    Otherwise, the network is pre-trained on the same amount of 10K segmentation data and then fine-tuned on the same amount of 200 matte data.
    The models are evaluated on P3M~\cite{(P3M)li2021privacy} dataset.
  }
  \label{tab:builtin}
\end{table}

\section{Conclusion and Future Direction}

In this paper, we introduced a new practical learning setting for human matting that utilizes both relatively easy-to-find coarse segmentation data and precise but expensive matte data.
Firstly, We observed that the segmentation data contains coarse-level human region information, and the model trained with matte data captures delicate details information mainly in the boundary region.
Based on the observation, the proposed Matte Label Blending method showed that guiding only informative knowledge from segmentation data and the model trained with matte data makes the matting model robust to natural images and precise boundary detail representation.
From the extensive experiments, we analyzed the effect of the amount of segmentation and matte data for human matting and verified the effectiveness of the proposed method.
Although we showed the segmentation data could be effectively used in human matting, we have a limit in the comfortable usability of an extensive pool of unlabeled images, such as web-crawling images.
Our promising future research direction will be applying pseudo-segmentation labels to utilize a huge amount of unlabeled images in the proposed framework.

\nocite{(pointwssis)kim2023devil}
\nocite{kim2022beyond}
\nocite{kim2021discriminative}
\nocite{lutz2018alphagan}
\nocite{(sinet)park2020sinet}
\nocite{(c3net)park2019extremec3net}
\nocite{(portraitnet)zhang2019portraitnet}
\nocite{papandreou2015weakly}
\nocite{lee2021anti}
\nocite{(wssod)yan2017weakly}
\nocite{(group_rcnn)zhang2022group}
\nocite{sun2021semantic}
\nocite{wei2021improved}
\nocite{lin2021real}
\nocite{yu2021high}
\nocite{sengupta2020background}
\nocite{li2020natural}
\nocite{hou2019context}
\nocite{cai2019disentangled}
\nocite{sun2021deep}
\nocite{lin2022robust}
\nocite{liu2021towards}
\nocite{dai2022boosting}
\nocite{ijcai2021-111}
\nocite{cheng2021deep} 
\nocite{zhong2021highly} 
\nocite{forte2020f}
\nocite{seong2022one}

{
    \small
    \bibliographystyle{ieeenat_fullname}
    \bibliography{ms}
}

\clearpage
\appendix
\renewcommand{\thesection}{\Alph{section}}
\section*{Appendix}

\section{Additional Analysis}

\subsection{Network architecture} 
\begin{figure*}[t]
    \centering
    \includegraphics[width=\linewidth]{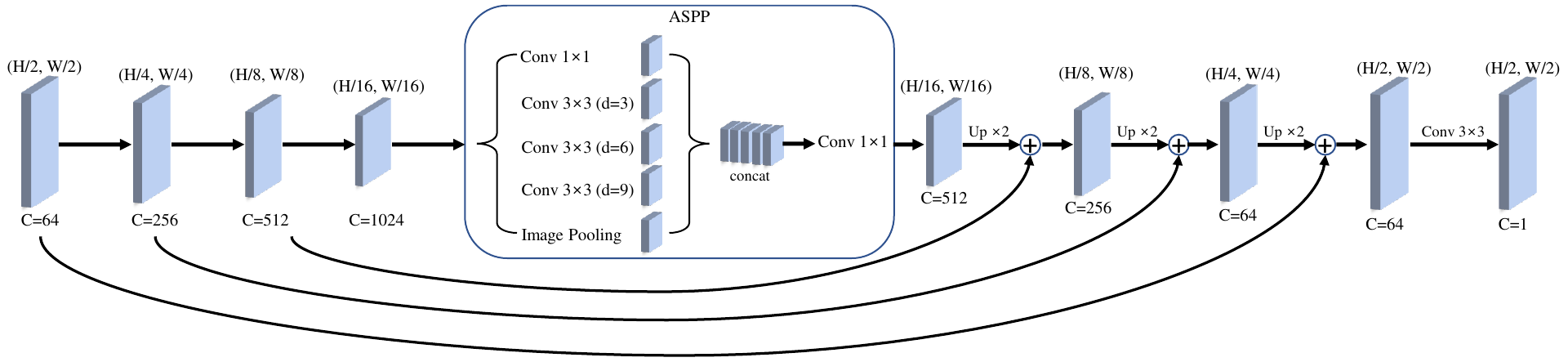}
    \caption{\textbf{Illustration of the network architecture we used.} We adopt the U-Net~\cite{(unet)ronneberger2015u} based architecture with the ResNet~\cite{(resnet)he2016deep} backbone network and ASPP module~\cite{(aspp)chen2017deeplab}.
    \texttt{Conv3$\times$3 (d=3)} means using a convolutional neural network with the kernel size of 3${\times}$3 and the dilation rate of 3.
    \texttt{Up$\times$2} means applying a bilinear upsampling layer.
    }
    \label{fig:network_architecture}
\end{figure*}

As mentioned in the experiment section, we used the U-Net~\cite{(unet)ronneberger2015u} structure with ResNet~\cite{(resnet)he2016deep} backbone network.
In \figurename~\ref{fig:network_architecture}, we provide an illustration of our network architecture.
Note that we remove the max-pooling layer in the stem part of the ResNet, and the architecture is highly straightforward and simple.

\begin{table*}[t]
  \centering
  \begin{adjustbox}{max width=\linewidth}
  \begin{tabular}{cc|cc|cc|cc|cc|cc|cc|cc|cc}
    \toprule
        & & \multicolumn{12}{c}{Natural Images} & \multicolumn{4}{|c}{Synthetic Images} \\
    \cline{3-18} \addlinespace[0.5ex]
        & & \multicolumn{4}{c|}{PPM~\cite{(PPM)ke2022modnet}} & \multicolumn{4}{c|}{P3M~\cite{(P3M)li2021privacy}} & \multicolumn{4}{c|}{RWP~\cite{(RWP)(MGMatte)yu2021mask}} & \multicolumn{4}{c}{Distinctions-646~\cite{(distinctions)(HAttMatting)qiao2020attention}} \\
    \hline
    \# seg & \# mat & \multicolumn{2}{c|}{Whole Region} & \multicolumn{2}{c|}{Boundary Region} & \multicolumn{2}{c|}{Whole Region} & \multicolumn{2}{c|}{Boundary Region} & \multicolumn{2}{c|}{Whole Region} & \multicolumn{2}{c|}{Boundary Region} & \multicolumn{2}{c|}{Whole Region} & \multicolumn{2}{c}{Boundary Region} \\
    \cline{3-18} \addlinespace[0.5ex]
           & & MSE $\downarrow$ & SAD $\downarrow$ & MSE $\downarrow$ & SAD $\downarrow$ & MSE $\downarrow$ & SAD $\downarrow$ & MSE $\downarrow$ & SAD $\downarrow$ & MSE $\downarrow$ & SAD $\downarrow$ & MSE $\downarrow$ & SAD $\downarrow $& MSE $\downarrow$ & SAD $\downarrow$ & MSE $\downarrow$ & SAD $\downarrow$ \\
    \midrule
    2K  & 0 & 5.129 & 154.675 & 68.699 & 65.668 & 6.907 & 29.991 & 50.727 & 15.329 & 17.943 & 50.777 & 95.872 & 28.608 & 11.562 & 99.361 & 55.793 & 46.424 \\
    5K  & 0 & 4.676 & 137.027 & 67.019 & 65.649 & 6.771 & 29.534 & 50.401 & 14.347 & 15.805 & 49.369 & 85.372 & 27.676 & 9.438 & 92.097 & 55.769 & 46.407 \\
    10K  & 0 & 4.375 & 124.083 & 66.326 & 64.557 & 6.676 & 28.158 & 48.088 & 14.077 & 15.279 & 48.191 & 84.883 & 28.969 & 9.125 & 90.853 & 53.630 & 45.329 \\
    20K  & 0 & 4.222 & 114.015 & 63.351 & 64.356 & 6.539 & 26.152 & 45.760 & 13.822 & 14.185 & 47.283 & 79.485 & 28.041 & 8.265 & 89.945 & 51.904 & 44.767 \\
    50K  & 0 & 4.130 & 102.522 & 62.662 & 64.186 & 6.493 & 26.113 & 43.031 & 13.577 & 14.072 & 46.108 & 78.211 & 27.845 & 7.001 & 80.785 & 47.563 & 43.558 \\
    \midrule
    0  & 50 & 30.781 & 532.989 & 99.097 & 78.266 & 50.220 & 137.725 & 93.092 & 21.302 & 161.199 & 180.452 & 204.548 & 43.815 & 15.584 & 99.183 & 53.696 & 47.369 \\
    2K  & 50 & 3.682 & 82.504 & 55.228 & 56.894 & 4.994 & 21.080 & 32.558 & 10.867 & 15.435 & 44.122 & 82.191 & 25.885 & 9.222 & 68.998 & 48.033 & 41.664 \\
    5K  & 50 & 3.380 & 81.419 & 52.964 & 55.520 & 4.764 & 20.824 & 31.519 & 10.723 & 13.678 & 42.349 & 74.699 & 24.994 & 7.397 & 58.224 & 44.578 & 40.306 \\
    10K  & 50 & 3.351 & 78.560 & 52.651 & 54.914 & 4.629 & 19.768 & 30.698 & 10.578 & 12.342 & 41.685 & 70.383 & 24.982 & 6.734 & 55.813 & 44.084 & 40.088 \\
    20K  & 50 & 2.980 & 74.714 & 51.526 & 53.537 & 4.593 & 19.132 & 29.871 & 10.499 & 11.891 & 41.277 & 67.690 & 24.855 & 5.808 & 55.055 & 42.805 & 39.580 \\
    50K  & 50 & 2.783 & 71.655 & 50.211 & 52.907 & 4.512 & 18.764 & 28.368 & 10.141 & 11.590 & 40.995 & 66.759 & 24.746 & 6.304 & 52.233 & 41.897 & 37.967 \\
    \midrule
    0  & 200 & 28.778 & 379.350 & 81.467 & 74.051 & 38.322 & 125.955 & 71.315 & 19.607 & 147.368 & 174.427 & 199.475 & 43.181 & 12.986 & 67.594 & 42.418 & 38.423 \\
    2K  & 200 & 3.321 & 77.307 & 51.109 & 54.613 & 4.743 & 19.943 & 30.573 & 10.423 & 13.960 & 35.479 & 71.251 & 25.117 & 8.739 & 58.641 & 44.689 & 40.136 \\
    5K  & 200 & 2.944 & 74.930 & 50.548 & 54.057 & 4.569 & 18.909 & 29.781 & 10.208 & 12.899 & 33.820 & 70.103 & 24.939 & 7.007 & 54.526 & 44.076 & 39.143 \\
    10K  & 200 & 2.815 & 72.908 & 50.240 & 53.624 & 4.472 & 18.392 & 28.837 & 10.168 & 11.974 & 32.871 & 66.965 & 24.905 & 5.601 & 54.049 & 40.366 & 38.530 \\
    20K  & 200 & 2.810 & 72.801 & 48.821 & 52.648 & 4.408 & 18.114 & 28.016 & 10.079 & 11.609 & 32.567 & 66.367 & 24.726 & 5.314 & 49.867 & 40.302 & 37.971 \\
    50K  & 200 & 2.636 & 70.763 & 48.512 & 52.112 & 4.303 & 17.755 & 27.900 & 10.027 & 11.496 & 31.173 & 66.044 & 24.589 & 5.438 & 49.512 & 39.985 & 37.144 \\
    \midrule
    0  & 500 & 25.513 & 358.315 & 77.480 & 65.381 & 36.656 & 97.700 & 69.166 & 16.868 & 134.276 & 172.330 & 182.532 & 40.766 & 11.852 & 46.528 & 40.290 & 35.524 \\
    2K  & 500 & 3.027 & 74.792 & 49.926 & 53.972 & 4.466 & 17.443 & 27.981 & 10.128 & 13.585 & 35.452 & 69.758 & 25.099 & 6.090 & 48.846 & 39.481 & 37.218 \\
    5K  & 500 & 2.902 & 73.733 & 48.846 & 52.937 & 4.260 & 16.975 & 27.691 & 9.991 & 12.870 & 33.499 & 67.908 & 24.900 & 5.255 & 48.753 & 39.117 & 36.762 \\
    10K  & 500 & 2.803 & 72.889 & 48.402 & 52.550 & 4.198 & 16.766 & 27.136 & 9.703 & 11.850 & 32.512 & 65.216 & 24.641 & 5.148 & 47.151 & 38.955 & 36.539 \\
    20K  & 500 & 2.740 & 71.639 & 48.111 & 52.075 & 4.073 & 16.681 & 25.541 & 9.635 & 11.480 & 32.417 & 65.016 & 24.573 & 5.043 & 46.847 & 38.095 & 36.320 \\
    50K  & 500 & 2.629 & 68.218 & 47.961 & 51.970 & 3.965 & 16.287 & 25.880 & 9.629 & 11.240 & 30.868 & 64.692 & 24.561 & 4.707 & 46.422 & 37.293 & 35.635 \\
    \midrule
    0  & 1000 & 22.102 & 278.938 & 74.582 & 62.741 & 36.123 & 92.371 & 62.685 & 15.423 & 125.071 & 162.275 & 181.543 & 40.163 & 4.598 & 45.086 & 37.086 & 35.213 \\
    2K  & 1000 & 3.050 & 71.727 & 49.598 & 53.074 & 4.316 & 16.402 & 27.052 & 9.967 & 13.461 & 35.382 & 69.004 & 25.062 & 5.131 & 47.592 & 39.291 & 36.588 \\
    5K  & 1000 & 2.844 & 70.688 & 48.650 & 51.962 & 4.074 & 16.371 & 26.943 & 9.836 & 12.848 & 32.988 & 64.773 & 24.786 & 4.860 & 47.473 & 37.692 & 35.889 \\
    10K  & 1000 & 2.746 & 70.187 & 48.370 & 51.928 & 3.979 & 16.002 & 25.837 & 9.517 & 11.789 & 31.406 & 64.149 & 24.532 & 4.805 & 45.857 & 37.265 & 35.865 \\
    20K  & 1000 & 2.700 & 68.380 & 48.060 & 51.918 & 3.766 & 15.974 & 24.663 & 9.511 & 11.379 & 31.311 & 63.536 & 24.525 & 4.795 & 45.494 & 37.230 & 35.618 \\
    50K  & 1000 & 2.601 & 67.634 & 47.813 & 51.553 & 3.695 & 15.562 & 24.556 & 9.455 & 11.164 & 30.734 & 63.085 & 24.310 & 4.670 & 45.368 & 37.172 & 35.612 \\
    \midrule
    0  & 2000 & 20.950 & 224.514 & 62.836 & 54.350 & 33.823 & 86.223 & 57.339 & 14.822 & 115.815 & 140.533 & 171.560 & 39.019 & 4.163 & 42.115 & 34.568 & 34.568 \\
    2K  & 2000 & 2.999 & 69.131 & 48.994 & 51.501 & 4.076 & 16.283 & 25.130 & 9.840 & 13.424 & 35.309 & 65.878 & 24.997 & 5.127 & 46.936 & 37.674 & 35.953 \\
    5K  & 2000 & 2.790 & 67.960 & 48.232 & 51.312 & 3.949 & 15.836 & 24.030 & 9.654 & 12.430 & 32.831 & 62.515 & 24.683 & 4.682 & 46.478 & 37.218 & 35.756 \\
    10K  & 2000 & 2.635 & 67.876 & 47.886 & 51.451 & 3.760 & 15.823 & 23.906 & 9.412 & 11.740 & 31.215 & 61.568 & 24.511 & 4.674 & 45.535 & 36.826 & 35.372 \\
    20K  & 2000 & 2.600 & 67.350 & 47.781 & 51.071 & 3.634 & 14.757 & 22.853 & 9.404 & 11.286 & 30.624 & 60.714 & 24.489 & 4.371 & 44.567 & 36.000 & 35.322 \\
    50K  & 2000 & 2.580 & 66.081 & 47.457 & 50.963 & 3.594 & 14.516 & 22.831 & 9.301 & 11.067 & 30.025 & 60.580 & 24.164 & 4.234 & 42.929 & 34.866 & 34.755 \\
    \bottomrule
  \end{tabular}
  \end{adjustbox}
  \caption{
    The whole numerical scores of our model on three validation sets consisting of natural images (PPM~\cite{(PPM)ke2022modnet}, P3M~\cite{(P3M)li2021privacy}, and RWP~\cite{(RWP)(MGMatte)yu2021mask}) and one synthetic validation set (Distinctions-646~\cite{(distinctions)(HAttMatting)qiao2020attention}) according to the amounts of segmentation and matte data.}
  \label{tab:whole_results}
\end{table*}
\begin{figure*}[t]
    \centering
    \includegraphics[width=0.75\linewidth]{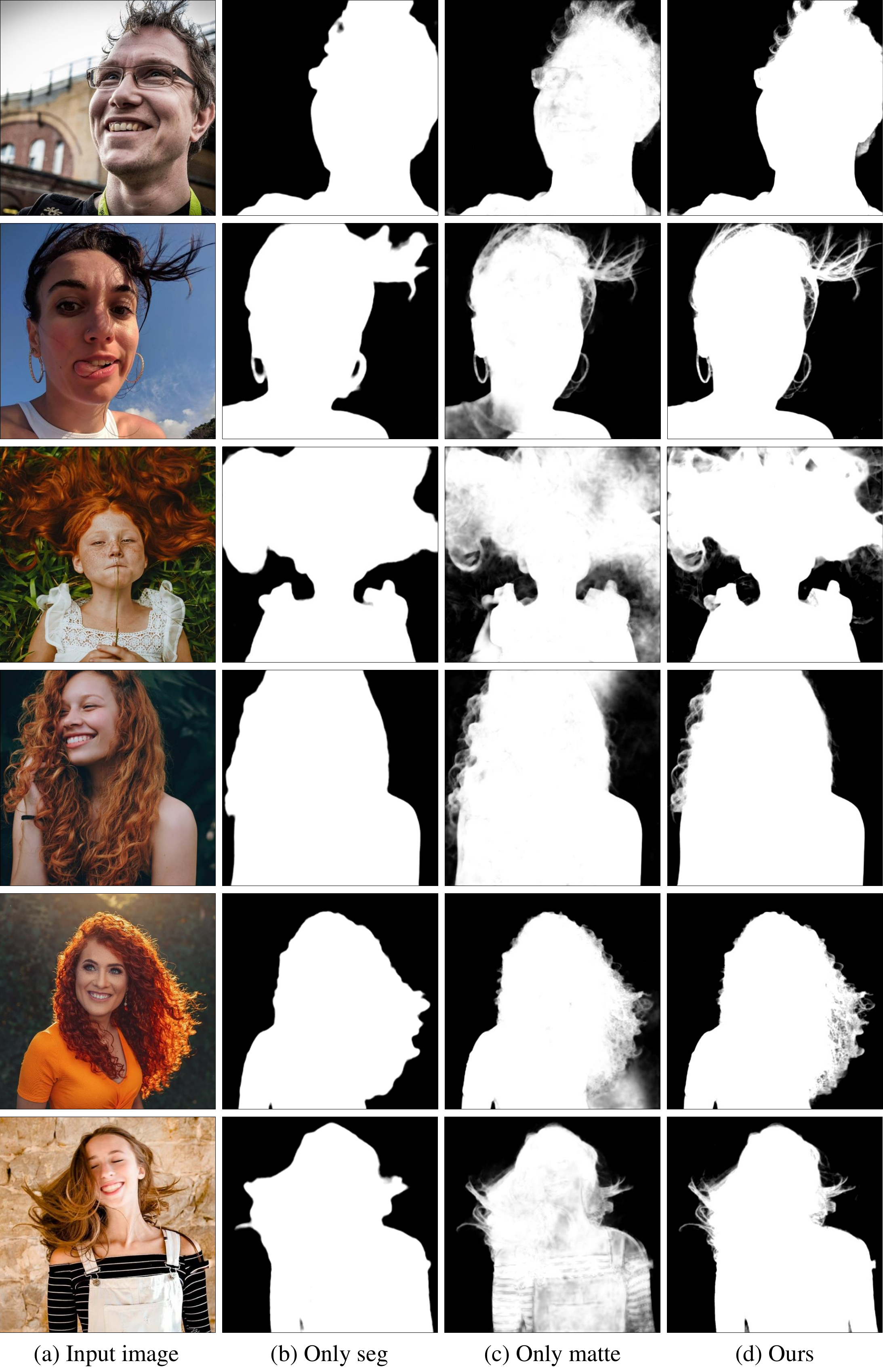}
    \caption{
    \textbf{Additional qualitative comparisons for the models}: the model trained with (b) only segmentation data, (c) only matte data, (d) and both segmentation and matte data with our training method. 
    }
    \label{fig:sample_appendix}
\end{figure*}

\subsection{Detailed quantitative results} In Figures \red{5--8}, we provided the quantitative scores of the matting model according to the amounts of segmentation and matte data in the form of a graph. Here, we provide all numerical scores including MSE and SAD metrics in Table~\ref{tab:whole_results}.

\subsection{Additional Qualitative results}
We provide more additional qualitative results in Figure~\ref{fig:sample_appendix}; compared to the models trained with only segmentation data or matte data, the model with the proposed method shows more robust results on natural images with finer boundary details.


\end{document}